\documentclass[journal]{IEEEtran}
\ifCLASSINFOpdf
\else
\fi

\usepackage{booktabs}
\usepackage{amsfonts}
\usepackage{graphicx}
\usepackage{multirow}
\usepackage{algorithm}
\usepackage{algpseudocode}
\usepackage{amsmath}
\usepackage{caption2}
\newtheorem{remark}{Remark}
\usepackage{color}

\begin{document}
%
\title{An Iterative Co-Training Transductive Framework for Zero Shot Learning}
%
%
%

\author{Bo Liu, Lihua Hu, Qiulei Dong, and Zhanyi Hu
\thanks{B. Liu and Z. Hu are with the National Laboratory of Pattern Recognition, Institute of Automation, Chinese Academy of Sciences, Beijing 100190, China, and also with the School of Future Technology, University of Chinese Academy of Sciences, Beijing 100049, China (e-mail: liubo2017@ia.ac.cn; huzy@nlpr.ia.ac.cn).}
\thanks{L. Hu is with the School of Computer Science and Technology, Taiyuan University of Science and Technology, Taiyuan 030024, China (e-mail: sxtyhlh@126.com).}
\thanks{Q. Dong is with the National Laboratory of Pattern Recognition, Institute of Automation, Chinese Academy of Sciences, Beijing 100190, China, also with the School of Artificial Intelligence, University of Chinese Academy of Sciences, Beijing 100049, China, and the Center for Excellence in Brain Science and Intelligence Technology, Chinese Academy of Sciences, Beijing 100190, China (e-mail: qldong@nlpr.ia.ac.cn).
\textit{(Corresponding author: Qiulei Dong.)}}}

%
%

\markboth{}%
{Shell \MakeLowercase{\textit{et al.}}: Bare Demo of IEEEtran.cls for IEEE Journals}
%



\maketitle

\begin{abstract}
In zero-shot learning (ZSL) community, it is generally recognized that transductive learning performs better than inductive one as the unseen-class samples are also used in its training stage. How to generate pseudo labels for unseen-class samples and how to use such usually noisy pseudo labels are two critical issues in transductive learning. In this work, we introduce an iterative co-training framework which contains two different base ZSL models and an exchanging module. At each iteration, the two different ZSL models are co-trained to separately predict pseudo labels for the unseen-class samples, and the exchanging module exchanges the predicted pseudo labels, then the exchanged pseudo-labeled samples are added into the training sets for the next iteration. By such, our framework can gradually boost the ZSL performance by fully exploiting the potential complementarity of the two models' classification capabilities. In addition, our co-training framework is also applied to the generalized ZSL (GZSL), in which a semantic-guided OOD detector is proposed to pick out the most likely unseen-class samples before class-level classification to alleviate the bias problem in GZSL. Extensive experiments on three benchmarks show that our proposed methods could significantly outperform about $31$ state-of-the-art ones.
\end{abstract}


%
\IEEEpeerreviewmaketitle

\section{Introduction}
\IEEEPARstart{R}{ecently}, deep learning methods have tremendously boosted the performance of object recognition. However, the success of deep learning methods heavily relies on a large number of labeled samples, which in some cases is an unrealistic demand for either labeling a large-scale dataset is expensive or new object categories are created everyday. To overcome these limitations, zero-shot learning (ZSL)~\cite{Lampert14DAP} has attracted increasing attention in the fields of machine learning and computer vision, which aims to recognize those classes (often named as unseen classes) whose samples lack labeling at the model training stage. In essence, ZSL is to transfer the learned knowledge from seen classes to unseen classes by learning an appropriate visual-semantic mapping between visual features and semantic features. Usually, the visual features are extracted by a convolutional neural network (CNN)~\cite{He16resnet} and semantic features are of attributes~\cite{Lampert14DAP}, word vectors~\cite{Frome13DeViSE}, and text descriptions~\cite{qiao2016ZSLNS}.

Most existing ZSL methods focused on learning a discriminative embedding space to establish the visual-semantic mapping. At the training stage, they projected the visual features and semantic features into a common embedding space, for instance, a semantic feature space~\cite{akata2015sje,xian2016LATEM,Meng2019ZeroShotLV,Rahman2018AUA} or a visual feature space~\cite{zhang2017DEM,Changpinyo17EXEM,EPGN2020,Gan2016LearningAE} or an intermediate feature space~\cite{Changpinyo16SYNC,Li2020AJL,jiang2019TCN,Gan2015ExploringSI}, by a linear or nonlinear mapping trained with the seen-class data. At the testing stage, the testing visual features and semantic features were projected into the learned embedding space and then the visual features were classified based on feature similarity. Despite the success of the embedding based methods, they generally suffer from two problems: 1) the domain shift problem that the learned visual-semantic mapping from the seen-class domain is not suitable for the unseen-class domain; 2) the bias problem in the generalized ZSL (GZSL) setting that the learned model is prone to recognize unseen classes as seen classes. More recently, the generative ZSL methods have received more attention for tackling the bias problem in GZSL. The generative methods~\cite{liu2020APNet,keshari2020OCDZSL,Vyas2020LrGAN,Long2018ZeroShotLU,Zhang2020APZ} proceeded by firstly generating many fake unseen-class samples conditioned on their corresponding semantic features via a conditional generative model trained with the seen-class data, then training a classifier with the synthetic unseen-class samples to classify real testing samples. Essentially speaking, the generative methods establish an one-to-many semantic-to-visual mapping between visual features and semantic features. Since their mappings are learned only from the seen-class data as done in the embedding based methods, they also suffer from the domain shift problem.

This work is to address the domain shift problem in ZSL and the bias problem in GZSL, by firstly proposing an iterative co-training transductive framework for ZSL (ICoT-ZSL). ICoT-ZSL consists of two base ZSL models and an exchanging module. At each iteration, the two base models are co-trained to separately predict pseudo labels for the unseen-class samples, and the exchanging module is to exchange the pseudo-labeled samples and add them to the training sets for the next iteration. This framework is able to fully exploit the potential complementarity of the base models' classification capabilities and gradually increases the final ZSL performance. Our proposed iterative co-training framework could also be easily extended to the multi-model cases. In this work, we extend this framework to three kinds of 3-model frameworks. In addition, we also adapt the proposed ICoT-ZSL framework to GZSL. To alleviate the bias problem in GZSL, we propose a novel semantic-guided OOD detector to pick out unseen-class samples from the compound unseen-class and seen-class samples before class-level classification. In sum, our main contributions include:
\begin{itemize}
	\item A general iterative co-training transductive framework, ICoT-ZSL, is introduced for ZSL. Its key advantage is to exploit the complementarity of different base ZSL models to gradually increase the ZSL performance. To our best knowledge, this is the first attempt in the literature to co-train two base models under the transductive setting to alleviate the domain shift problem.
	\item Our ICoT-ZSL is also adapted to GZSL, where a novel semantic-guided OOD detector is proposed to alleviate the bias problem in GZSL by pre-classifying the testing samples into the most probably unseen-class samples, which are then classified by ICoT-ZSL, and the remaining ones, which are classified by ICoT-GZSL.
	\item Extensive experimental results largely validate our proposed framework. Our proposed methods significantly outperform $31$ state-of-the-art ones on three public benchmarks with two data splits. In addition, a comprehensive analysis on these results is also provided to demonstrate the effectiveness of the key components of the framework.
\end{itemize}

The remaining of this paper is organized as follows. Firstly, we review some related works in Section~\ref{related}. Secondly, we elaborate the proposed methods in Section~\ref{methodology}, where the ICoT-ZSL framework and the semantic-guided OOD detector are discussed. In Section~\ref{experiments}, the experimental results and an in-depth discussion on the components in the ICoT-ZSL framework and semantic-guided OOD detector are provided. Finally, we conclude the paper and outline some future works in Section~\ref{conclusion}.

\section{Related Work}
\label{related}

\subsection{Inductive Zero-Shot Learning}
As a pioneering work, Lampert et al.~\cite{Lampert14DAP} proposed a two-stage method for ZSL, where a probabilistic classifier was firstly learned for predicting probability of each attribute for each image, then the image was classified by a Bayesian classifier based on the probabilities of attributes. Recently, most ZSL methods focused on learning a discriminative embedding space to establish a mapping between visual features and semantic features in an end-to-end manner. According to the specific embedding space, these methods could be roughly divided into three categories. The first one~\cite{Frome13DeViSE,akata2015sje,romera2015ESZSL,xian2016LATEM,socher2013CMT,Xie_2019_AREN,Zhu19SGML,YeG19PREN,Gan2016RecognizingAA} learned a visual-to-semantic mapping which projected visual features into the semantic feature space and then classified each visual feature according to the distances between the projected feature and all semantic features. The second one~\cite{zhang2017DEM,Changpinyo17EXEM,Luo2018ZeroShotLV,EPGN2020} learned a semantic-to-visual prediction function which projected semantic features into the visual feature space to predict the corresponding visual prototypes, then a visual feature could be classified based on the distances between the visual feature and the predicted visual prototypes. The third one~\cite{Li2020AJL,Changpinyo16SYNC,jiang2019TCN} projected visual features and semantic features into a common intermediate feature space and classified the testing visual features according to a distance metric. More recently, generative ZSL methods have received much attention due to their better performances. The generative methods employed a conditional generative adversarial network (GAN)~\cite{Xian18FCLSWGAN,elhoseiny2017ZSLPP,AFRNetLiuDH20,li2019LiGAN,paul2019SABR,YuL19SGL,Jia2020DeepUE} or a conditional variational autoencoder (VAE)~\cite{Mishra2018CVAE,xian2019f-VAEGAN,schonfeld2019CADA-VAE,Gao2020ZeroVAEGANGU} to generate many fake unseen-class samples conditioned on the corresponding semantic features, then such fake samples with their corresponding labels were used to train a classifier for classifying the real unseen-class ones.

\subsection{Transductive Zero-Shot Learning}
Recently, transductive ZSL~\cite{Yu2018TransductiveZL,xian2019f-VAEGAN,wu2020SDGN,Zhang2020DeepTN,Zhang2020TowardsED,bo2021hardness} has received much attention, where the unlabeled unseen-class data are used together with the seen-class data at the training stage. Fu et al.~\cite{Fu15TMV} proposed a graph based method where visual features and semantic features were projected into a multi-view embedding space and then a hyper-graph was constructed using the unlabeled data for label propagation. Kodirov et al.~\cite{Kodirov15uda} considered the ZSL problem as a domain adaptation problem and proposed the UDA method which used unsupervised domain adaption with sparse coding to alleviate the domain shift problem. Guo et al.~\cite{Guo16sms} proposed the SMS method which leaned a shared model space on seen-class and unseen-class data. Song et al.~\cite{Song18QFSL} proposed the QFSL method where the model was trained to predict large softmax probabilities of unseen classes on the unseen-class data. Both Ye et al.~\cite{YeG19PREN} and Li et al.~\cite{li2019GXE} proposed an iterative training based method where a fixed number of pseudo-labeled unseen-class data were selected from the unseen-class dataset for network re-training. More recently, the generative models~\cite{Narayan2020TF-VAEGAN,paul2019SABR} were applied to transductive ZSL, where a shared generator was used to generate seen-class and unseen-class visual features conditioned on their corresponding semantic features, and two discriminators were employed to adversarially train the shared generator with the labeled seen-class data and unlabeled unseen-class data respectively.

\subsection{Domain-Aware Generalized Zero-Shot Learning}
Recently, many methods have been proposed to alleviate the bias problem in GZSL. For example, \cite{Das2019ZeroshotIR, Oreshkin2020CLARELCV,ChaoCGS16,Liu2018DCN} employed a calibration strategy where a predefined constant was subtracted from the prediction probabilities of seen classes to reduce the bias towards seen classes. Others resorted to the out-of-distribution (OOD) detection one. In~\cite{socher2013CMT}, a Gaussian mixture model was used to estimate the probability of an input sample being an OOD one. In~\cite{mandal2019CE-WGAN-OD}, a conditional GAN was utilized to generate fake OOD samples and a two-class classifier was trained with real seen-class samples and fake OOD samples to classify the inputs into seen and unseen classes. In our proposed method, we also employ an OOD detection module. However, our OOD detector is different from the existing works in 1) semantic features are used to guide OOD detection and 2) an iterative training strategy is proposed to iteratively train the OOD detector.

\subsection{Pseudo-Label Method}
Pseudo-label methods are widely used in the semi-supervised learning. Lee et al.~\cite{Lee2013PseudoLabel} proposed a pseudo-label based method where unlabeled data were assigned to the most confident class by a deep neural network and the network was re-trained with the pseudo-labeled data. Blum et al.~\cite{BlumM98cotrain} proposed a co-training algorithm which utilized two classifiers to provide pseudo labels and then added the pseudo-labeled samples into the training set, which were confidently predicted by at least one classifier. Zhou et al.~\cite{ZhouL05tritrain} proposed a tri-training algorithm which employed three classifiers to provide pseudo labels.

To our best knowledge, this work is the first attempt to co-train two models iteratively to tackle the ZSL task. In the next section, our proposed iterative co-training transductive framework will be elaborated.

\section{Methodology}
\label{methodology}

\subsection{ZSL, GZSL, and Transductive Setting}
At first, we introduce the definitions of ZSL, GZSL and their transductive settings. At the training stage, assume that we have a seen-class dataset $\mathcal{D}^{S}_{tr} = \{(x_{n}, y_{n})\}_{n=1}^{N}$ , where $x_{n}$ is the $n$-th visual feature and $y_{n}$ is the label of $x_{n}$ which belongs to the seen-class label set $Y^{S}$, and $N$ is the number of samples in $\mathcal{D}^{S}_{tr}$. The semantic feature set $\mathcal{E}=\{e_{y} \mid y \in Y \}$ is also given at the training stage, where $Y$ is the total class label set which includes both the seen-class label set $Y^{S}$ and the unseen-class label set $Y^{U}$. Note that $Y^{S}$ is disjoint with $Y^{U}$. At the testing stage, assume that we have a testing unseen-class dataset $\mathcal{D}^{U}$ and a testing seen-class dataset $\mathcal{D}^{S}$. For ZSL, the task is to learn a mapping $F: \mathcal{D}^{U} \to Y^{U}$ with the training seen-class dataset $\mathcal{D}^{S}_{tr}$ and the semantic feature set $\mathcal{E}$. For GZSL, the task is to learn a mapping $F: \mathcal{D}^{S} \cup \mathcal{D}^{U} \to Y$. In the transductive settings, we assume that the testing unseen-class dataset $\mathcal{D}^{U}$ is available for training in a ZSL task, and the testing unseen-class dataset $\mathcal{D}^{U}$ and testing seen-class dataset $\mathcal{D}^{S}$ are both available for training in a GZSL task.

\begin{figure}
	\centering
	\includegraphics[width=1\linewidth]{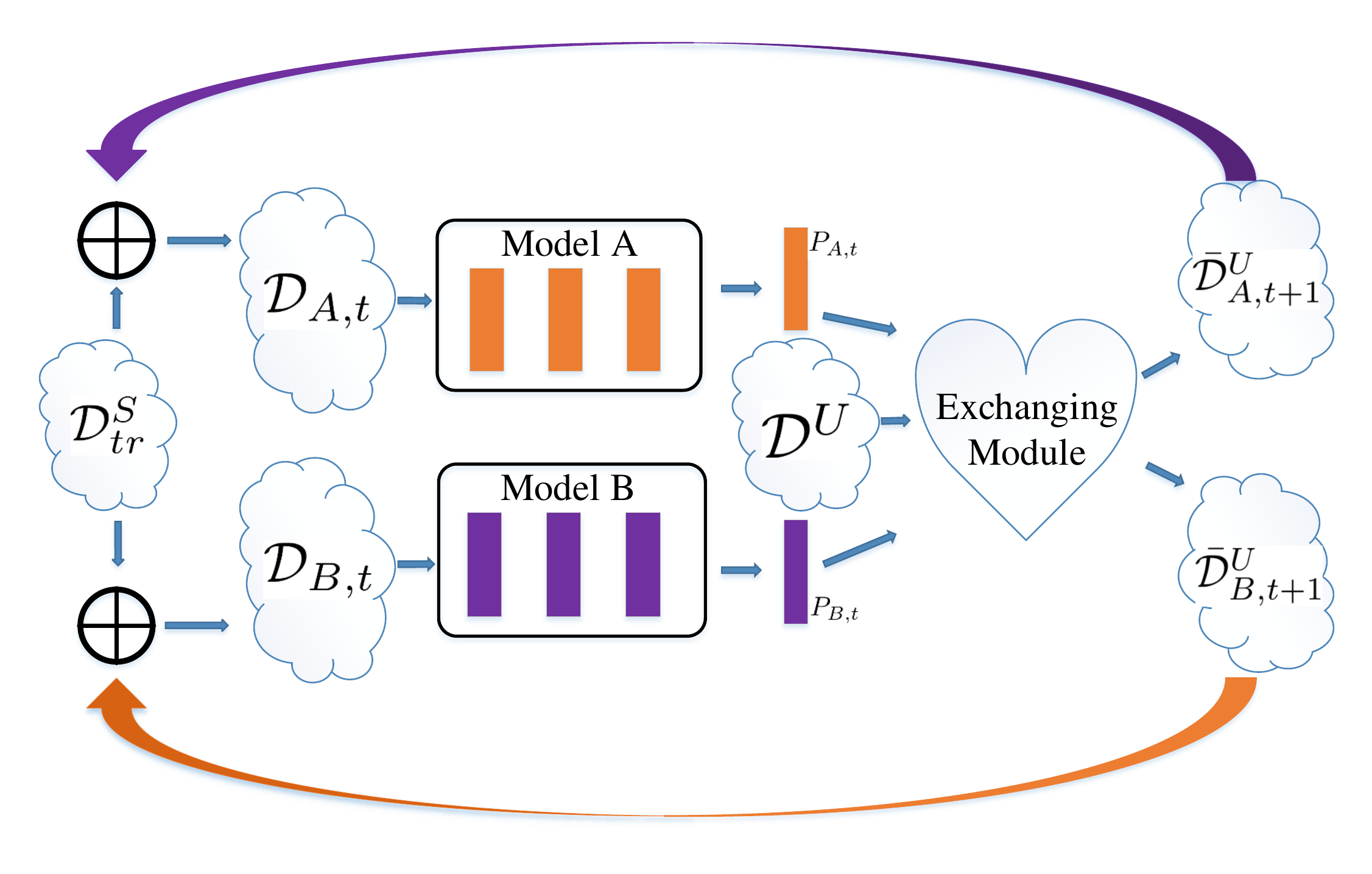}
	\caption{The proposed ICoT-ZSL framework.}
	\label{fig1}
\end{figure}

In this section, we firstly propose an iterative co-training transductive framework for ZSL, called ICoT-ZSL, which consists of two base ZSL models and an exchanging module, as shown in Fig.~\ref{fig1}. Then, we generalize the proposed framework to GZSL. To alleviate the bias problem in GZSL, we employ a two-stage method as shown in Fig.~\ref{fig2}, where a semantic-guided OOD detector is proposed to differentiate the unseen-class samples from the seen-class samples before class-level classification.

\subsection{An Iterative Co-Training Transductive Framework for ZSL}

\subsubsection{The Overall Iterative Co-Training Transductive Framework}
Several existing works~\cite{YeG19PREN,li2019GXE} employed a single-model self-training strategy to make use of unlabeled unseen-class samples for alleviating the domain-shift problem in ZSL. However, the single-model iterative training is easy to be entrapped into a `local minima' because the single model is iteratively trained with the unseen-class samples pseudo-labeled by itself at each iteration. Considering that two models would make diverse predictions on unseen-class samples, co-training two models is expected to alleviate the `local minima' problem. Here we introduce an iterative co-training transductive framework for ZSL, called ICoT-ZSL. As shown in Fig.~\ref{fig1}, the ICoT-ZSL framework consists of two base ZSL models and an exchanging module. The core idea is to iteratively co-train the two base models via the exchanging module to help them learn more appropriate visual-semantic mappings simultaneously. Specifically, we assume that the whole training process includes $T$ iterations, and the two base models are denoted as Model A and Model B respectively. At the $t$-th ($t \in \{1, 2, \cdots, T\}$) iteration, Model A and Model B are firstly trained with their corresponding training sets $\mathcal{D}_{A,t}$ and $\mathcal{D}_{B,t}$, each of which includes not only the labeled seen-class dataset $\mathcal{D}^{S}_{tr}$ but also a subset of pseudo-labeled unseen-class data, i.e. $\mathcal{\bar{D}}^{U}_{A,t}$ or $\mathcal{\bar{D}}^{U}_{B,t}$. Then two sets of probabilistic predictions are made on the whole unseen-class dataset by the trained Model A and Model B, which are denoted as $P_{A,t}$ and $P_{B,t}$ respectively. Based on these predictions $P_{A,t}$ and $P_{B,t}$, we could easily obtain two corresponding pseudo label sets, denoted as $Y_{A,t}$ and $Y_{B,t}$ respectively. Next, two new subsets of unseen-class samples, denoted as $\mathcal{\bar{D}}^{U}_{A,t+1}$ and $\mathcal{\bar{D}}^{U}_{B,t+1}$, are selected for Model A and Model B by the exchanging module from the two pseudo-labeled unseen-class datasets respectively, where $\mathcal{\bar{D}}^{U}_{A,t+1}$ is selected based on $Y_{B,t}$ while $\mathcal{\bar{D}}^{U}_{B,t+1}$ is selected based on $Y_{A,t}$. The exchange of pseudo labels between two base models is helpful for the two base models to fully exploit the potential complementarity of their classification capabilities. Finally, the training sets for Model A and Model B are updated by combining the two selected new unseen-class subsets $\mathcal{\bar{D}}^{U}_{A,t+1}$ and $\mathcal{\bar{D}}^{U}_{B,t+1}$ with the seen-class dataset $\mathcal{D}^{S}_{tr}$ for re-training Model A and Model B at the $t+1$-th iteration, i.e. $\mathcal{D}_{A,t+1}=\mathcal{D}^{S}_{tr}+\mathcal{\bar{D}}^{U}_{A,t+1}$ and $\mathcal{D}_{B,t+1}=\mathcal{D}^{S}_{tr}+\mathcal{\bar{D}}^{U}_{B,t+1}$. After $T$ iterations, for a given testing sample, the output of the ICoT-ZSL framework is obtained by computing the weighted sum of the probabilistic predictions from Model A and Model B as follows:
\begin{equation}
	y = \arg \max_{y \in Y^{U}} (\alpha p_{A,T} + (1-\alpha) p_{B,T})
	\label{eq1}
\end{equation}
where $p_{A,T}$ and $p_{B,T}$ are the probabilistic predictions on the testing sample by Model A and Model B respectively. The complete procedure of the proposed ICoT-ZSL framework is summarized in Algorithm~\ref{alg1}.

Note that our ICoT-ZSL is a general transductive framework, it can in fact accommodate any existing inductive ZSL models. Here are two guidelines for the selection of the two base models: Firstly, they should have relatively good performances in ZSL, otherwise the subsequent iterations would be less effective. This criterion is quantitatively measured by average per-class accuracy ($ACC$) of base models in this work; Secondly, they should possess some complementary ZSL capabilities. Considering the general law in computational neuroscience that structure determines the function, the two base models should ideally have different network architectures, or their scatter matrices under some representative datasets should be diverse. To quantitatively measure this criterion, we propose the average per-class ratio of different predictions ($APR$) in our current implementation. Specifically, $APR$ is computed by firstly computing the ratio of the number of different predictions made by two base models on each unseen class to the total number of samples belonging to corresponding unseen class and then computing the mean of per-unseen-class ratios. Note also that we empirically find the cross-model exchanging scheme performs better, we do not have any theoretical basis to exclude other exchanging modes, for example, if one base model performs much better than the other, no cross-model exchanging is needed, and only the better model's pseudo-labeled samples is used for the next iteration.

\subsubsection{Incremental Learning Scheme}
Considering that the two base models have relatively lower accuracies at the initial stages of the iterative training process, i.e. the noise in the predicted pseudo labels is relatively higher, we propose an incremental learning scheme to select the unseen-class samples instead of an one-off learning scheme as done in~\cite{YeG19PREN,li2019GXE}. More specifically, at the $t$-th iteration, $[M*t/T]$ unseen-class samples are selected from the whole unseen-class dataset for constructing the new unseen-class set ($\mathcal{\bar{D}}^{U}_{A,t+1}$ or $\mathcal{\bar{D}}^{U}_{B,t+1}$) by randomly sampling an equal number of samples from each pseudo label class with replacement, where $[\cdot]$ is a round-down function and $M$ is the number of samples in the unseen-class dataset $\mathcal{D}^{U}$, and `with replacement' is to prevent some classes from overwhelming others and to handle the insufficiency of the number of samples in some classes. In other words, a small number of unseen-class samples are selected at the initial stages when the accuracies of base models are relatively low. As the iterative training goes on, the accuracies of the base models are progressively improved, and more pseudo-labeled samples are selected for model re-training.
\begin{remark}
Here we would point out that although our proposed ICoT-ZSL framework also includes an iterative training process as done in~\cite{YeG19PREN,li2019GXE}, the proposed ICoT-ZSL framework is considerably different from them in three aspects: 1) The proposed ICoT-ZSL framework co-trains two base models via an exchanging module to learn their visual-semantic mappings while only one network is designed to learn the visual-semantic mapping in~\cite{YeG19PREN,li2019GXE}; 2) The proposed ICoT-ZSL is a general framework, and it can accommodate any existing inductive ZSL models; 3) The proposed ICoT-ZSL framework uses unlabeled unseen-class samples in a manner quite different from those in~\cite{YeG19PREN,li2019GXE}.
\end{remark}

\begin{algorithm}[t]
	\caption{ICoT-ZSL}
	\begin{algorithmic}[1]
		\Require
		$\mathcal{D}^{S}_{tr}$, $\mathcal{D}^{U}$, $\mathcal{E}$;
		\Ensure
		Predicted labels $Y_{T}$ on $\mathcal{D}^{U}$;
		\State Initialization: train Model A and Model B with $\mathcal{D}^{S}_{tr}$ and $\mathcal{E}$, make predictions on $\mathcal{D}^{U}$, and construct $\mathcal{D}_{A,1}$, $\mathcal{D}_{B,1}$ via the exchanging module;
		\For{t=1 to T}
		\State Train Model A and Model B with $\mathcal{D}_{A,t}$, $\mathcal{D}_{B,t}$, and $\mathcal{E}$;
		\State Make predictions $P_{A,t}$ and $P_{B,t}$ on $\mathcal{D}^{U}$ with the trained models;
		\State Select unseen-class subsets $\mathcal{\bar{D}}^{U}_{A,t+1}$ and $\mathcal{\bar{D}}^{U}_{B,t+1}$, and update $\mathcal{D}_{A,t+1}$ and $\mathcal{D}_{B,t+1}$ via the exchanging module;
		\EndFor
		\State Predict labels $Y_{T}$ on $\mathcal{D}^{U}$ according to function in (\ref{eq1});\\
		\Return $Y_{T}$;
	\end{algorithmic}
	\label{alg1}
\end{algorithm}

\subsubsection{Multi-Model Case}
Our proposed iterative co-training transductive framework could be easily extended to the multi-model cases by simply 1) employing more models in the framework and 2) altering the exchanging module accordingly. Clearly, the performance of the multi-model framework is affected by many factors, such as the number of used base models, the performances of individual base models, as well as the used exchanging modules. It is beyond the scope of this work to give a systematic investigation on such factors. Our understanding is that basically, when the available base ZSL models have comparable performances, it is not the number of the base models used in the framework that counts, but the complementarity of their ZSL capabilities.

To verify this claim, here we design three different 3-model frameworks by extending the proposed ICoT-ZSL framework. The first one includes two Model A (initialized differently) and one Model B, denoted as 2A+B; the second one includes one Model A and two Model B (initialized differently), denoted as A+2B; the third one includes one Model A, one Model B, and one Model C, denoted as A+B+C. In addition, we employ two pseudo-label exchanging schemes to implement the exchanging module in each of the 3-model frameworks. The first one is a cyclic exchanging module where the three base models constitute a cycle and the pseudo labels are propagated between the three base models along this cycle. For instance, A1$\rightarrow$A2$\rightarrow$B$\rightarrow$A1 in the 2A+B framework. The second one is an agreement-based exchanging module where the pseudo-labeled samples and corresponding pseudo labels of one model are selected from those samples on which the other two models have the same predictions. We conduct some experiments with the three 3-model frameworks and two exchanging modules, and the corresponding experimental results and some discussions are presented in Section \uppercase\expandafter{\romannumeral4}-G.

\begin{remark}
Note that in both 2A+B and A+2B, a very limited amount of complementary information is expected to be additionally added compared to the 2-model ICoT-ZSL framework (i.e. A+B) since the added third base model (i.e. Model A in 2A+B, or Model B in A+2B) has the same architecture with one of the two base models in ICoT-ZSL. By such designs, we hope that the effect of the number of the used base models with respect to their ZSL complementarity could be revealed. In addition, since the base model's performance also affects the final performance of the iterative framework, we use the same two base models (i.e. Model A and Model B) in these two 3-model frameworks as in the 2-model ICoT-ZSL framework so that we could fairly compare the performances of the two 3-model frameworks with that of the 2-model framework.

In contrast to 2A+B and A+2B, A+B+C usually introduces more complementary information since the added Model C has different architecture from both Model A and Model B. To avoid the possibility that the good performance of A+B+C comes mainly from a much higher performance of Model C with respect to those of Model A and Model B, a Model C with plain performance is used in our current implementation.
\end{remark}

\subsection{ICoT-GZSL and Semantic-OOD for GZSL}
We adapt the proposed ICoT-ZSL framework to GZSL, and the adapted framework is denoted as ICoT-GZSL. The ICoT-GZSL framework is different from the ICoT-ZSL framework in the following two aspects: 1) two base GZSL models are employed for co-training in the ICoT-GZSL framework and 2) the testing unseen-class dataset and testing seen-class dataset are compound at the training stage, consequently, the pseudo labels are predicted in the total class space in the ICoT-GZSL framework. The overall procedure of ICoT-GZSL is summarized in Algorithm~\ref{alg2}. Since the base models perform GZSL tasks in the ICoT-GZSL framework, the bias problem also exists in the ICoT-GZSL framework. To address the bias problem, we propose a semantic-guided OOD detection method (named as Semantic-OOD) to pick out the most likely unseen-class samples from the compound testing unseen-class and testing seen-class samples before class-level classification. In the following, we first present the proposed Semantic-OOD method. Then we briefly introduce the overall pipeline to perform GZSL based on Semantic-OOD.
\begin{algorithm}[t]
	\caption{ICoT-GZSL}
	\begin{algorithmic}[1]
		\Require
		$\mathcal{D}^{S}_{tr}$, $\mathcal{D}^{U} \cup \mathcal{D}^{S}$, $\mathcal{E}$;
		\Ensure
		Predicted labels $Y_{T}$ on $\mathcal{D}^{U} \cup \mathcal{D}^{S}$;
		\State Initialization: train Model A and Model B with $\mathcal{D}^{S}_{tr}$ and $\mathcal{E}$, make predictions on $\mathcal{D}^{U} \cup \mathcal{D}^{S}$, and construct $\mathcal{D}_{A,1}$, $\mathcal{D}_{B,1}$ via the exchanging module;
		\For{t=1 to T}
		\State Train Model A and Model B with $\mathcal{D}_{A,t}$, $\mathcal{D}_{B,t}$, and $\mathcal{E}$;
		\State Make predictions $P_{A,t}$ and $P_{B,t}$ on $\mathcal{D}^{U} \cup \mathcal{D}^{S}$ with the trained models in the GZSL setting;
		\State Select unseen-class subsets $\mathcal{\bar{D}}^{U}_{A,t+1}$ and $\mathcal{\bar{D}}^{U}_{B,t+1}$, and update $\mathcal{D}_{A,t+1}$ and $\mathcal{D}_{B,t+1}$ via the exchanging module;
		\EndFor
		\State Predict labels $Y_{T}$ on $\mathcal{D}^{U} \cup \mathcal{D}^{S}$;\\
		\Return $Y_{T}$;
	\end{algorithmic}
	\label{alg2}
\end{algorithm}

\subsubsection{Semantic-OOD}
In the proposed Semantic-OOD method, the seen-class data are considered as in-distribution (ID) data while the unseen-class data are considered as out-of-distribution (OOD) data. At the training stage, besides the labeled seen-class dataset, the compound testing seen-class and testing unseen-class datasets are also available. Hence, the proposed OOD detector is in fact learned in a transductive manner. Before elaborating on the proposed OOD detector, let us consider the following ideal situation: suppose we had the access to a set of unseen-class data $\mathcal{\bar{D}}^{U}$, which are representative of the real unseen-class data, then the OOD detection task would have become relatively easy since we could train an OOD detector as follows:
\begin{equation}
	\min_{F} E_{(x,y) \in \mathcal{D}^{S}_{tr}}[ \mathcal{C}(F(x), y)] + E_{\grave{x} \in \mathcal{\bar{D}}^{U}}[ KL(F(\grave{x}) || U)]
	\label{eq7}
\end{equation}
where $F(\cdot)$ is the OOD detector, $\mathcal{C}$ is a cross entropy loss function, $KL$ represents the KL divergence, $U$ is an uniform distribution, and $\mathcal{D}^{S}_{tr}$ is the labeled seen-class dataset. After training, the model could be able to output high-entropy predictions on the unseen-class data and low-entropy predictions on the seen-class data, and the OOD detection could be accomplished by setting an entropy threshold.

However, since the given testing seen-class and testing unseen-class datasets are compound in our task, the ideal unseen-class set $\mathcal{\bar{D}}^{U}$ is not directly available. Hence, the problem turns into how to select a representative unseen-class set from the given compound seen-class and unseen-class data to simulate $\mathcal{\bar{D}}^{U}$. Here, we propose a semantic-guided method to select an unseen-class set to simulate $\mathcal{\bar{D}}^{U}$. For clarity, we denote the ideal unseen-class set by $\mathcal{\bar{D}}^{U}$ and denote the simulated one by $\mathcal{\hat{D}}^{U}$. Specifically, we first learn a semantic-guided classifier with the labeled seen-class dataset $\mathcal{D}^{S}_{tr}$ as follows:
\begin{equation}
	\min_{F_{sc}} E_{(x,y) \in \mathcal{D}^{S}_{tr}} [\mathcal{C}( \mathcal{S}(\langle F_{sc}(x),\mathcal{E}_{Y^{S}} \rangle), y)]
	\label{eq8}
\end{equation}
where $F_{sc}(\cdot)$ is the semantic-guided classifier, $\mathcal{E}_{Y^{S}}$ is the seen-class semantic feature set, $\langle \cdot \rangle$ is the dot product, $\mathcal{S}$ and $\mathcal{C}$ are the softmax function and the cross entropy loss function respectively. After training, given a testing sample $x$, it is classified by projecting $F_{sc}(x)$ into the semantic feature space which includes not only the seen classes but also the unseen classes, i.e. $\mathcal{E}_{Y^{S}}$ is replaced by $\mathcal{E}_{Y}$. Since the classifier is learned from only seen-class data, the prediction probabilities on seen classes should generally be relatively higher for all the testing samples, and if a testing sample is predicted with higher probabilities on unseen classes, then this sample is most likely to be a real unseen-class one. Based on this idea, we could select a simulated unseen-class set $\mathcal{\hat{D}}^{U}$. Finally, we train an OOD detector with the simulated unseen-class set $\mathcal{\hat{D}}^{U}$ according to the loss function in (\ref{eq7}). Since $\mathcal{\hat{D}}^{U}$ is provided by a semantic-guided classifier, we name this method as Semantic-OOD.

For comparison, we also propose another method to select the simulated unseen-class set $\mathcal{\hat{D}}^{U}$. Specifically, we first train a base OOD detector using a baseline method~\cite{Hendrycks17Baseline} as follows:
\begin{equation}
	\min_{F_{c}} E_{(x,y) \in \mathcal{D}^{S}_{tr}} [\mathcal{C}(F_{c}(x), y)]
	\label{eq9}
\end{equation}
where $F_{c}(\cdot)$ is the base OOD detector. After training, the base OOD detector detects the unseen-class samples according to the order of prediction confidences. Specifically, it selects $L$ samples with the lowest prediction confidences as the simulated unseen-class set $\mathcal{\hat{D}}^{U}$. Finally, we re-train a new OOD detector according to the loss function in (\ref{eq7}) as done in Semantic-OOD. Since $\mathcal{\hat{D}}^{U}$ is provided by a base OOD detector, we name this method as Iter-OOD.

\subsubsection{Semantic-OOD based GZSL}
\begin{figure}
	\centering
	\includegraphics[width=0.9\linewidth]{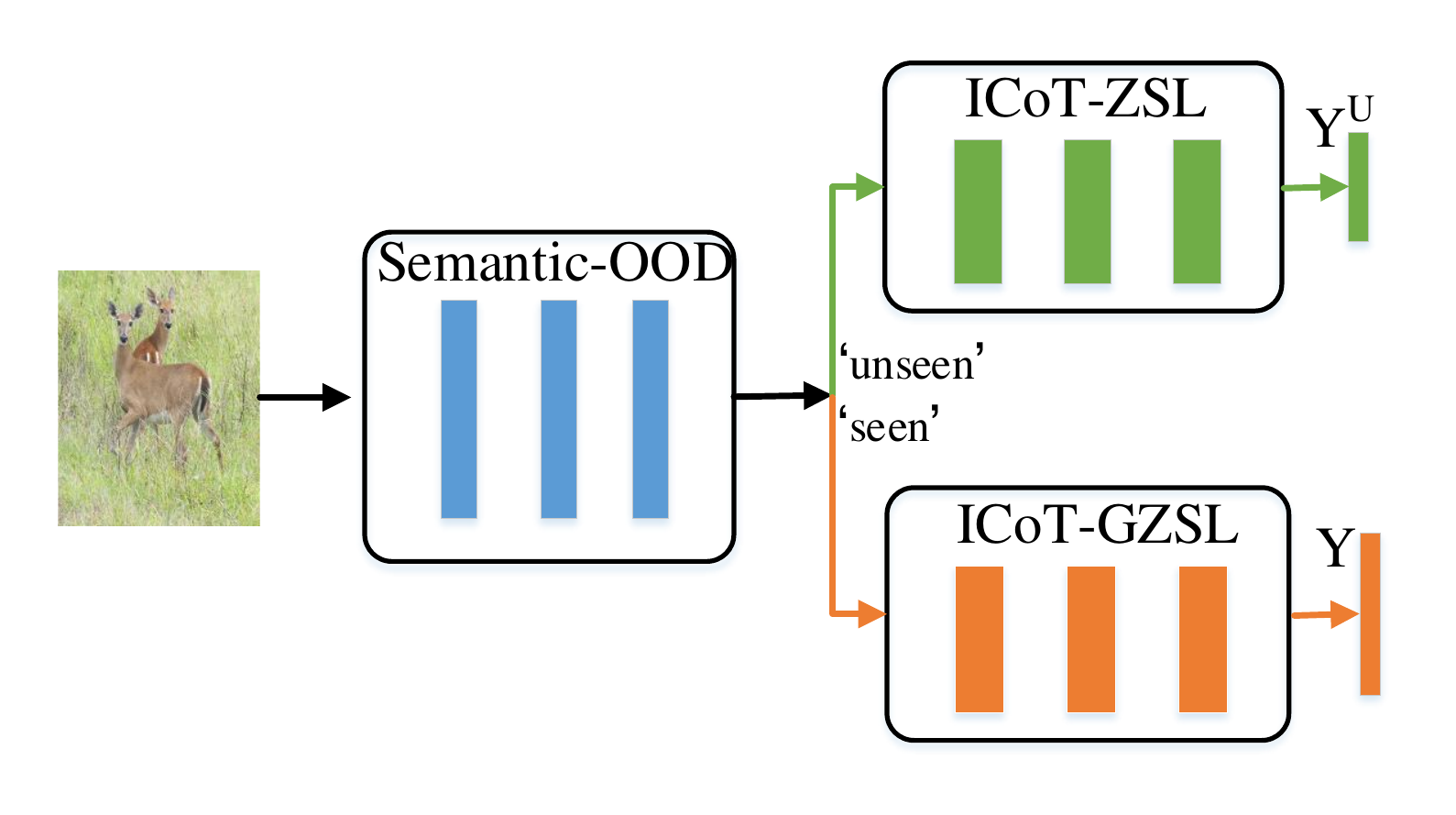}
	\caption{The two-stage ICoT-ZSL-SOD for GZSL.}
	\label{fig2}
\end{figure}

In order to tackle the bias problem in GZSL, as shown in Fig.~\ref{fig2}, we employ a two-stage method to perform GZSL, named as ICoT-ZSL-SOD, where we firstly pick out the most likely unseen-class samples from the compound seen-class and unseen-class samples with the proposed Semantic-OOD, and then the picked unseen-class samples are classified by the proposed ICoT-ZSL while the remaining ones are classified by the proposed ICoT-GZSL. In some OOD detection based GZSL methods~\cite{dvbe2020,atzmon2019COSMO}, a seen-class classifier is employed to classify the remaining ones. Here we employ a GZSL method to classify the remaining ones because 1) limited by the accuracy of Semantic-OOD, the remaining ones include not only seen-class samples but also some unseen-class samples. It is impossible to correctly classify these unseen-class samples if we employ a seen-class classifier, while it is possible if a GZSL method is employed; 2) as the seen-class accuracy achieved by a GZSL model is generally high, the seen-class accuracy is not hampered by the usage of GZSL model.

\section{Experiment}
\label{experiments}
\subsection{Datasets and Comparative Methods}
We evaluate the proposed methods on three public benchmarks, i.e. AWA2 (renewed Animals with Attributes~\cite{Xian17Comprehensive}), CUB (Caltech USCD Birds-2011~\cite{WahCUB_200_2011}), SUN (SUN attributes~\cite{patterson2012sun}). AWA2 is an animal dataset which contains 37,322 images from 50 animal classes and each class is annotated with 85 attributes. CUB is a fine-grained bird dataset which includes 11,788 images from 200 birds species and 312 attributes are annotated for each species. SUN is a fine-grained scene dataset which includes 14,340 images belonging to 717 scene classes and each class is annotated with 102 attributes. As done in most existing methods, we employ the class-level attributes as semantic features, and the visual features of the three datasets which are extracted using the ImageNet1000 pre-trained ResNet101~\cite{He16resnet} are used as model inputs in the proposed methods. The detailed statistics about the three datasets are summarized in Table~\ref{tab1}.
\begin{table}[t]
	\setlength{\abovecaptionskip}{0cm}	
	\setlength{\belowcaptionskip}{0.1cm}
	\caption{Statistics about AWA2, CUB and SUN. Visual: the dimensionality of visual features, Semantic: the dimensionality of semantic features, S: the number of seen classes, U: the number of unseen classes.}
	\centering
	\begin{tabular}{cccccccc}
		\toprule
		Dataset&  Number&  Visual&  Semantic&  \multicolumn{2}{c}{SS}&  \multicolumn{2}{c}{PS} \\
		\cmidrule(lr){5-6} \cmidrule(lr){7-8}
		& & & & S& U& S& U\\
		\cmidrule{1-8}
		AWA2&  	 37,322&	  		2048&	  85&   40&	  10&    40&   10\\
		CUB&  	 11,788&	  		2048&	  312&  150&  50&    150&  50\\
		SUN&  	 14,340&	  		2048&	  102&  645&  72&    645&  72\\
		\bottomrule
	\end{tabular}
	\label{tab1}
\end{table}

The proposed methods are compared with $31$ state-of-the-art ZSL methods, including $14$  inductive ZSL methods: SJE~\cite{akata2015sje}, ESZSL~\cite{romera2015ESZSL}, SAE~\cite{Kodirov17SAE}, DEM~\cite{zhang2017DEM}, f-CLSWGAN~\cite{Xian18FCLSWGAN}, ABP~\cite{zhu2019ABP}, TCN~\cite{jiang2019TCN}, DASCN~\cite{NiZ019dascn}, OCD-GZSL~\cite{keshari2020OCDZSL}, APNet~\cite{liu2020APNet}, DE-VAE~\cite{Ma2020DE-VAE}, LsrGAN~\cite{Vyas2020LrGAN}, DVBE~\cite{dvbe2020}, DAZLE~\cite{huynh2020DAZLE} and $17$ transductive ZSL methods: UDA~\cite{Kodirov15uda}, TMV~\cite{Fu15TMV}, SMS~\cite{Guo16sms}, ALE-tran~\cite{Akata16ALE}, GFZSL~\cite{verma2017GFZSL}, DSRL~\cite{Ye2017DSRL}, QFSL~\cite{Song18QFSL}, GMN~\cite{sariyildiz2019GMN}, f-VAEGAN-D2~\cite{xian2019f-VAEGAN}, GXE~\cite{li2019GXE}, SABR-T~\cite{paul2019SABR}, PREN~\cite{YeG19PREN}, VSC~\cite{wan2019vsc}, ADA~\cite{Khare2020ADA}, DTN~\cite{Zhang2020DeepTN}, EDE~\cite{Zhang2020TowardsED}, Zero-VAE-GAN~\cite{Gao2020ZeroVAEGANGU}.

\subsection{Evaluation Protocol}
The standard split (SS) and the proposed split (PS)~\cite{Xian17Comprehensive} are two popular approaches to split seen/unseen classes used by the existing ZSL methods. The SS data split was widely used by the early ZSL methods, however, as some unseen classes in SS have been used for training by the ImageNet1000 pre-trained CNNs which are usually used to extract visual features by most existing ZSL methods, Xian et al.~\cite{Xian17Comprehensive} proposed the PS data split to compensate this problem. In both SS and PS data split, the numbers of seen classes in AWA2, CUB, SUN are $\{40,150,645\}$ respectively, and the remaining $\{10,50,72\}$ classes are regarded as unseen classes respectively. The details about the two data splits on the three datasets are reported in Table~\ref{tab1}. As done in most existing methods, we evaluate the proposed methods with both SS and PS data split in the conventional ZSL setting and evaluate them with the PS data split in the generalized ZSL setting. For the conventional ZSL, average per-class Top-1 accuracy ($ACC$) on unseen classes is usually adopted to evaluate the ZSL performance. For the generalized ZSL, the $ACC$ on unseen classes and that on seen classes are computed respectively, and then their harmonic mean $H$ is computed to evaluate the GZSL performance as:
\begin{equation}
	H = \frac{2 \times ACC_{seen} \times ACC_{unseen}}{ACC_{seen} + ACC_{unseen}}
	\label{eq10}
\end{equation}

\subsection{Implementation Details}
We employ two typical ZSL methods as the base models in the ICoT-ZSL framework. Model A employs a prototype predictor to predict visual prototypes with their corresponding semantic features and classifies a given unseen-class visual sample based on the distances between the sample and the predicted unseen-class visual prototypes, which is slightly modified from~\cite{zhang2017DEM}. Model B firstly generates many fake unseen-class visual features conditioned on their corresponding semantic features via a conditional WGAN and then trains a classifier with these synthetic samples, which is slightly modified from~\cite{Xian18FCLSWGAN}. Note that since the main concern of this paper is the ICoT-ZSL framework, we employ two base ZSL models with relatively plain performances compared with recent state-of-the-art methods. The iterative steps $T$ are empirically set as $\{10,7,9\}$ on AWA2, CUB, SUN respectively. For the training of Model A, the training epochs on AWA2, CUB, SUN are set as $\{5,30,30\}$ respectively and the learning rates and batch sizes on all the three datasets are set as $0.001$ and $128$ respectively. For the training of Model B, the training epochs and batch sizes on all the three datasets are set as $30$ and $256$ respectively, and the learning rates on AWA2, CUB, SUN are set as $\{0.00005,0.0001,0.0001\}$. In the extended A+B+C framework, a Model C which has a different architecture from both Model A and Model B is employed. In Model C, many fake unseen-class visual features are firstly generated conditioned on the semantic features via a conditional generator, and then a classifier is trained with these fake samples to classify real unseen-class ones. In the semantic-guided OOD detector, the model is implemented by a three-layer fully-connected neural network whose input-unit number and latent-unit number are $2048$ and $1600$ respectively, and the output-unit number is the number of corresponding seen classes. The model is trained by $50$ epochs with learning rate of $0.001$ and batch size of $256$ on all the three datasets. All the models are trained with Adam optimizer. The training parameters are summarized in Table~\ref{tab2}.
\begin{table*}[t]
	\setlength{\abovecaptionskip}{0cm}	
	\setlength{\belowcaptionskip}{0.1cm}
	\centering
	\caption{Configuration of training parameters.}
		\begin{tabular}{cccccccccc}
			\toprule
			Dataset& \multicolumn{3}{c}{Model A}& \multicolumn{3}{c}{Model B}& \multicolumn{3}{c}{Semantic-OOD}\\
			\cmidrule(lr){2-4} \cmidrule(lr){5-7} \cmidrule(lr){8-10}
			&	Epochs&	 LR&  BS&  Epochs&	 LR&  BS&    Epochs&  LR&  BS\\	
			\cmidrule{1-10}
			AWA2& 5&  0.001&  128&  30&  0.00005&  256&  50&  0.001&  256\\
			CUB&  30&  0.001&  128&  30&  0.0001&  256&  50&  0.001&  256\\
			SUN&  30&  0.001&  128&  30&  0.0001&  256&  50&  0.001&  256\\
			\bottomrule
		\end{tabular}
	\label{tab2}
\end{table*}

\subsection{Performance on Conventional ZSL}
Here we evaluate our proposed ICoT-ZSL framework in the conventional ZSL setting on AWA2, CUB and SUN with both SS and PS data splits and then compare them with $28$ existing methods. The corresponding results are reported in Table~\ref{tab5} where the methods marked by $\mathcal{I}$ are inductive ZSL methods while those marked by $\mathcal{T}$ are transductive ZSL methods. Those methods marked by * employed a fine-tuned CNN to extract visual features instead of the ImageNet1000 pre-trained CNN used by the others. The results of the comparative methods are cited from either the original papers or the public results~\cite{Xian17Comprehensive}. As seen from Table~\ref{tab5}, our proposed method significantly outperforms the comparative methods in most cases. Specifically, it achieves an improvement about $1.8\%$ on AWA2 with the SS data split, improvements about $3.4\%$ and $0.6\%$ on CUB with the SS and PS data split respectively, and improvements about $1.1\%$ and $3.0\%$ on SUN with the SS and PS data split respectively. Note that the second-ranking methods (i.e. VSC and SABR-T) both employ a fine-tuned CNN to extract visual features, which contributes notably to their high performances since the visual features extracted by the fine-tuned CNN are more discriminative than those extracted by a pre-trained CNN. For a fairer comparison, we also compare our proposed method only with those methods using the same visual features as inputs. We find that the improvements become more significant. For instance, the improvements on CUB with SS and PS become $7.9\%$ and $5.7\%$ respectively, and the improvement on SUN with PS becomes $3.7\%$. All these results demonstrate that our proposed method could effectively make use of unlabeled unseen-class data to learn more suitable visual-semantic mappings, resulting in a significantly superior performance.
\begin{table}[t]
	\setlength{\abovecaptionskip}{0cm}	
	\setlength{\belowcaptionskip}{0.1cm}
	\caption{Comparative results ($ACC$) in the conventional ZSL setting on AWA2, CUB, and SUN.}
	\centering
	\begin{tabular}{cccccccc}
		\toprule
		&Method& \multicolumn{2}{c}{AWA2}&  \multicolumn{2}{c}{CUB}&   \multicolumn{2}{c}{SUN} \\
		\cmidrule(lr){3-4} \cmidrule(lr){5-6} \cmidrule(lr){7-8}
		&& 			  SS&	  PS&		SS& 	PS&			SS& 	PS	  \\
		\cmidrule{1-8}
		\multirow{11}{*}{$\mathcal{I}$}
		&SJE~\cite{akata2015sje}&  	  				69.5&	  61.9&		55.3& 	53.9&	57.1& 	53.7\\
		&ESZSL~\cite{romera2015ESZSL}&  	  			75.6&	  58.6&		55.1& 	53.9&	57.3& 	54.5\\
		&SAE~\cite{Kodirov17SAE}&  	  				80.7&	  54.1&		33.4& 	33.3&	42.4& 	40.3\\
		&DEM~\cite{zhang2017DEM}&  	  				   -&	  67.1&		   -& 	51.7&	   -& 	61.9\\
		&f-CLSWGAN~\cite{Xian18FCLSWGAN}&  	  		   -&	     -&		   -& 	57.3&	   -& 	60.8\\
		&SABR~\cite{paul2019SABR}&  -&    65.2&        -&   63.9&      -&   62.8 \\
		&ABP~\cite{zhu2019ABP}&     -&    70.4&        -&   58.5&      -&   61.5 \\
		&TCN~\cite{jiang2019TCN}&   -&    71.2&        -&   59.5&      -&   61.5 \\	
		&OCD-GZSL~\cite{keshari2020OCDZSL}&  			   -&	  71.3&        -&   60.3&      -&   63.5\\
		&DE-VAE~\cite{Ma2020DE-VAE}&    			   -&	  69.3&        -&   63.1&      -&   64.0\\
		&LsrGAN~\cite{Vyas2020LrGAN}&    			   -&	     -&        -&   60.3&      -&   62.5\\
		\cmidrule{1-8}
		\multirow{18}{*}{$\mathcal{T}$}
		&UDA~\cite{Kodirov15uda}&  	  				      -&	     -&		39.5& 	   -&	   -& 	-\\
		&TMV~\cite{Fu15TMV}&  	  				      -&	     -&		51.2& 	   -&	61.4& 	-\\
		&SMS~\cite{Guo16sms}&  	  				      -&	     -&		59.2& 	   -&	60.5& 	-\\
		&f-VAEGAN~\cite{xian2019f-VAEGAN}&  	  	   -&	  \textbf{89.8}&		-& 	    64.6&	   -& 	64.3\\
		&GMN~\cite{sariyildiz2019GMN}&  	  			   		   -&	     -&		-& 	    64.6&	   -& 	64.3\\
		&SABR-T*~\cite{paul2019SABR}&  	  			       -&	  88.9&		-& 	    74.0&	   -& 	67.5\\
		&GXE~\cite{li2019GXE}&  	  			           -&	  83.2&		-& 	    61.3&	   -& 	63.5\\
		&ALE-tran~\cite{Akata16ALE} &  	  			   -&	  70.7&		-& 	    54.5&	   -& 	55.7\\
		&GFZSL~\cite{verma2017GFZSL}&  	  			       -&	  78.6&		-& 	    50.0&	   -& 	64.0\\
		&DSRL~\cite{Ye2017DSRL}&  	  			       -&	  72.8&		-& 	    48.7&	   -& 	56.8\\		
		&QFSL*~\cite{Song18QFSL}&  	  				84.8&	  79.7&	 69.7& 	    72.1&	61.7& 	58.3\\
		&PREN~\cite{YeG19PREN}&  	  				95.7&	  74.1&	  66.9& 	66.4&	63.3& 	62.9\\
		&VSC*~\cite{wan2019vsc}&  	  				    96.8&	  81.7&	  73.6& 	71.0&	66.2& 	62.2\\
		&ADA~\cite{Khare2020ADA}&  	  				       -&	  78.6&	     -& 	   -&	   -& 	65.5\\
		&DTN~\cite{Zhang2020DeepTN}&  	  				       -&	  -&	     -& 	   61.1&	   -& 	65.6\\
		&EDE~\cite{Zhang2020TowardsED}&  	  				       -&	  77.5&	     -& 	   67.8&	   -& 	61.6\\
		&Zero-VAEGAN~\cite{Gao2020ZeroVAEGANGU}&  	  			-&	  85.4&	     69.1& 	   68.9&	   68.4& 	66.8\\
		&ICoT-ZSL(Ours)& \textbf{98.6}& \textbf{89.8}& \textbf{77.0}& \textbf{74.6}& \textbf{69.5}& \textbf{70.5}\\
		\bottomrule
	\end{tabular}
	\label{tab5}
\end{table}

\subsection{Performance on Generalized ZSL}
Here we evaluate our proposed methods in the generalized ZSL setting on AWA2, CUB and SUN with the PS data split and compare them with $19$ existing methods. According to the existing transductive generalized ZSL (T-GZSL) methods, there are two data settings for T-GZSL. In the first setting, the testing unseen-class dataset is separately used at the training stage, but they are compound with the testing seen-class dataset at the testing stage, as done in~\cite{li2019GXE,paul2019SABR,wan2019vsc,xian2019f-VAEGAN,sariyildiz2019GMN}. In the second setting, the testing unseen-class dataset and testing seen-class dataset are compound at both training and testing stages, as done in~\cite{YeG19PREN,Akata16ALE,verma2017GFZSL,Ye2017DSRL}. Intuitively, the GZSL task in the first setting is easier since unlabeled samples are known to be unseen classes. For a fair comparison, we adapt our proposed framework for both data settings and compare them with the corresponding methods.

In the first setting, since the testing unseen-class dataset could be separately used for training, we propose a two-stage method to perform GZSL, denoted as ICoT-ZSL-OOD, where we first train an OOD detector using the seen-class dataset and the testing unseen-class dataset according to the loss function in (\ref{eq7}) for domain classification, and then a seen-class classifier and a ICoT-ZSL method are used to perform intra-domain classification. The results of our proposed two-stage method and corresponding comparative methods (marked by $\mathcal{T}_{1}$) are reported in Table~\ref{tab6}. As seen from Table~\ref{tab6}, our proposed method achieves significantly superior performances over the most recent state-of-the-art methods. Specifically, the improvements on AWA2, CUB, and SUN are $4.2\%$, $4.3\%$, and $3.7\%$ respectively, which demonstrate the effectiveness of our proposed ICoT-ZSL method and the two-stage method.

In the second setting, since the testing unseen-class dataset and testing seen-class dataset are compound at the training stage, we employ the proposed ICoT-ZSL-SOD method to perform GZSL tasks. The results of the ICoT-ZSL-SOD method and corresponding comparative methods (marked by $\mathcal{T}_{2}$) are reported in Table~\ref{tab6}. From Table~\ref{tab6}, we can see that 1) the performances achieved in the first data setting are generally better than in the second data setting. This is reasonable because the predicted pseudo labels of the unlabeled samples are more likely to be accurate in the first setting since the testing samples are assumed to be in the unseen-class space; 2) our proposed ICoT-ZSL-SOD method outperforms all the comparative methods in most cases. In particular, the improvements are $4.4\%$ and $7.3\%$ on AWA and CUB respectively. These significant improvements demonstrate that our proposed co-training transductive framework is effective by exploiting unlabeled samples to reduce the domain shift problem and our proposed Semantic-OOD method appears exceptionally good at differentiating the unseen-class samples from the seen-class samples. Besides, note that the seen-class accuracies and unseen-class accuracies of our proposed ICoT-ZSL-SOD method are more balanced than the competitors, hence a high $H$.
\begin{table*}[t]
	\setlength{\abovecaptionskip}{0cm}	
	\setlength{\belowcaptionskip}{0.1cm}
	\caption{Comparative results in the generalized ZSL setting on AWA2, CUB, and SUN. U: $ACC$ on unseen classes, S: $ACC$ on seen classes, and $H$: the harmonic mean of U and S.}
	\centering
	\begin{tabular}{ccccccccccc}
		\toprule
		&Method& \multicolumn{3}{c}{AWA2}&  \multicolumn{3}{c}{CUB}&   \multicolumn{3}{c}{SUN}\\
		\cmidrule(lr){3-5} \cmidrule(lr){6-8} \cmidrule(lr){9-11}
		& & 			  				U&	  S&	 H& 	U&	  S&	 H&		U&	  S&	 H	  \\
		\cmidrule{1-11}
		\multirow{7}{*}{$\mathcal{I}$}
		& DASCN~\cite{NiZ019dascn}&                  -& -& -& 45.9& 59.0& 51.6& 42.4& 38.5& 40.3 \\
        & DVBE~\cite{dvbe2020}&        		   63.6& 70.8& 67.0& 53.2& 60.2& 56.5& 45.0& 37.2& 40.7 \\
        & DAZLE~\cite{huynh2020DAZLE}&       		   60.3& 75.7& 67.1& 56.7& 59.6& 58.1& 52.3& 24.3& 33.2 \\
        & OCD-GZSL~\cite{keshari2020OCDZSL}&        	   59.5& 73.4& 65.7& 44.8& 59.9& 51.3& 44.8& 42.9& 43.8 \\
        & APNet~\cite{liu2020APNet}&       		   54.8& 83.9& 66.4& 48.1& 55.9& 51.7& 35.4& 40.6& 37.8 \\
        & DE-VAE~\cite{Ma2020DE-VAE}&      		   58.8& 78.9& 67.4& 52.5& 56.3& 54.3& 45.9& 36.9& 40.9 \\
        & LsrGAN~\cite{Vyas2020LrGAN}&                 -& -& -& 48.1& 59.1& 53.0& 44.8& 37.7& 40.9 \\
		\cmidrule{1-11}
		\multirow{7}{*}{$\mathcal{T}_{1}$}
		& f-VAEGAN-D2~\cite{xian2019f-VAEGAN}&    		 84.8& 88.6& 86.7& 61.4& 65.1& 63.2& 60.6& 41.9& 49.6 \\
		& GMN~\cite{sariyildiz2019GMN}&         				-& -& -& 60.2& 70.6& 65.0& 57.1& 40.7& 47.5 \\
		& SABR-T~\cite{paul2019SABR}&      			 79.7& 91.0& 85.0& 67.2& 73.7& 70.3& 58.8& 41.5& 48.6 \\
        & GXE~\cite{li2019GXE}&         			 80.2& 90.0& 84.8& 57.0& 68.7& 62.3& 45.4& 58.1& 51.0 \\
        & VSC~\cite{wan2019vsc}&  		  		 71.9& 88.2& 79.2& 33.1& 86.1& 47.9& 29.9& 62.9& 40.6  \\
        &Zero-VAEGAN~\cite{Gao2020ZeroVAEGANGU}&   70.2&    87.0&    77.6&   64.1&  57.9&  60.8&  53.1&  35.8& 42.8 \\
		&ICoT-ZSL-OOD(Ours)& 			 89.6& 92.3& \textbf{90.9}& 74.6& 74.6& \textbf{74.6}& 70.3& 44.8& \textbf{54.7}\\
		\cmidrule{1-11}		
		\multirow{7}{*}{$\mathcal{T}_{2}$}	
		&ALE-tran~\cite{Akata16ALE}& 			    12.6& 73.0& 21.5& 23.5& 45.1& 30.9& 19.9& 22.6& 21.2 \\
		&GFZSL~\cite{verma2017GFZSL}&     			   -&    -&    -& 24.9& 45.8& 32.2&    -&    -& - \\
        &DSRL~\cite{Ye2017DSRL}&        			   -&    -&    -& 17.3& 39.0& 24.0& 17.7& 25.0& 20.7 \\	
		&PREN~\cite{YeG19PREN}&  		  		 32.4& 88.6&  47.4&  35.2& 55.8&  43.1&	 35.4& 27.2& 30.8  \\
		&DTN~\cite{Zhang2020DeepTN}&     			   -&    -&    -&  42.6& 66.0& 51.8&    35.8&    38.7& 37.2 \\
		&EDE~\cite{Zhang2020TowardsED}&    68.4&    93.2&  78.9&  54.0& 62.9& 58.1&    47.2&    38.5&   \textbf{42.4}\\
		&ICoT-ZSL-SOD(Ours)& 	 84.8& 81.8& \textbf{83.3}& 66.6& 64.1& \textbf{65.4}& 50.4& 36.6& \textbf{42.4}\\
		\bottomrule
	\end{tabular}
	\label{tab6}
\end{table*}

\subsection{Performance on OOD Detection}
\begin{figure}
	\centering
	\includegraphics[width=1.0\linewidth]{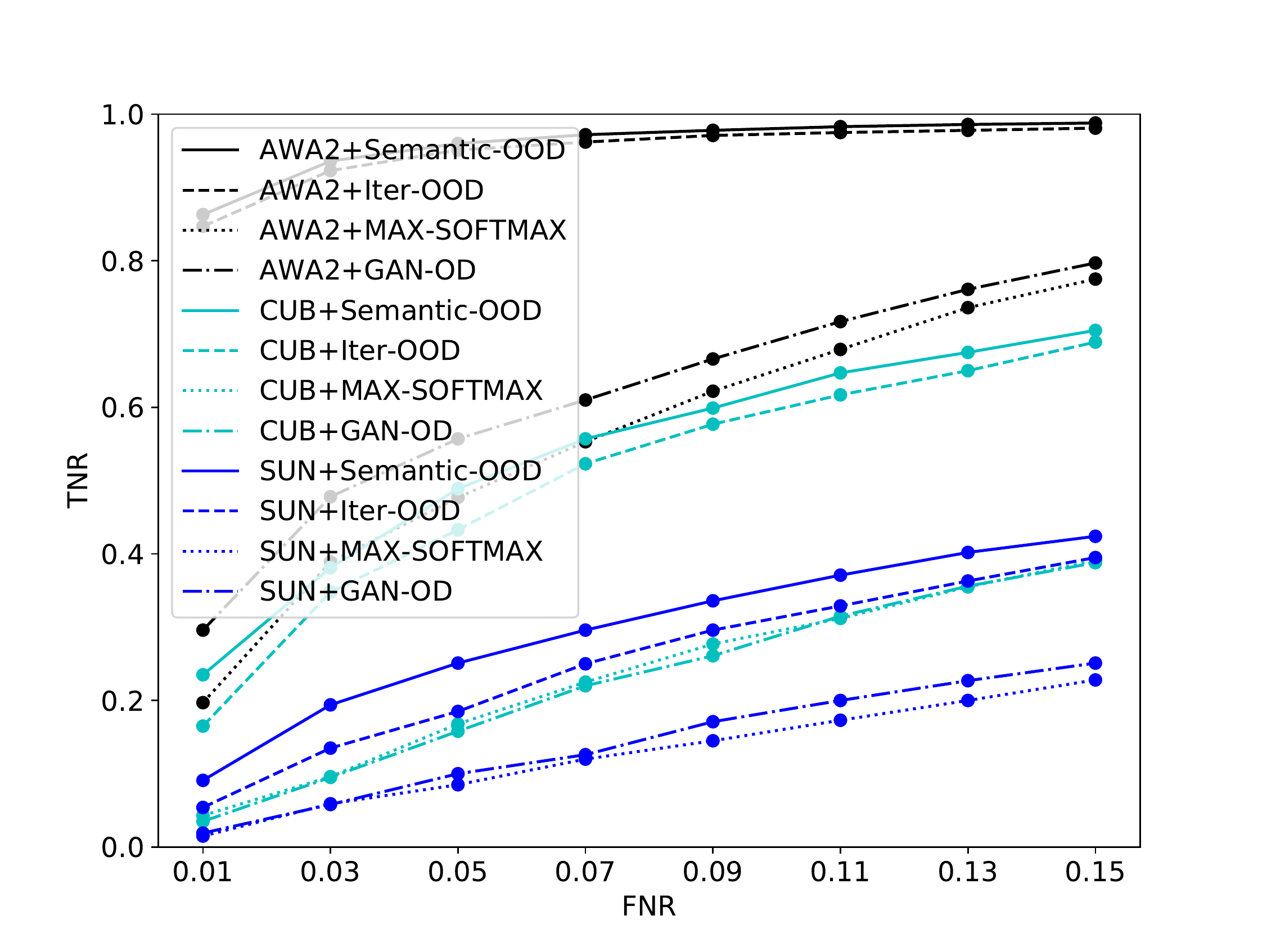}
	\caption{Comparative results of OOD detectors on AWA2, CUB, and SUN.}
	\label{fig4}
\end{figure}

Here we evaluate the performance of our proposed OOD detectors by performing OOD detection tasks on AWA2, CUB and SUN with the PS data split. Since OOD detection methods are usually not evaluated on the three datasets, we select two typical methods whose codes are available for comparison: 1) MAX-SOFTMAX~\cite{Hendrycks17Baseline} which trains a classifier with the seen-class dataset and then differentiates the seen and unseen class samples by the maximum of softmax probabilities; 2) GAN-OD~\cite{mandal2019CE-WGAN-OD} which generates many fake unseen-class samples conditioned on their corresponding semantic features via a WGAN model and then trains a seen/unseen classifier to perform OOD detection. Note that MAX-SOFTMAX and GAN-OD are both inductive methods. The performance of OOD detection is evaluated by True-Negative Rate (TNR) under a series of False-Negative Rate (FNR), where unseen-class data are considered as the negative and the FNRs are set as $\{0.01, 0.03, 0.05, 0.07, 0.09, 0.11, 0.13, 0.15\}$. Fig.~\ref{fig4} shows the corresponding TNRs under different FNRs and Table~\ref{tab9} reports the average TNR, where the results of MAX-SOFTMAX and GAN-OD are obtained with public codes. As seen from Fig.~\ref{fig4}, firstly, we find that our proposed Semantic-OOD and Iter-OOD both significantly outperform MAX-SOFTMAX and GAN-OD on all the three datasets. This demonstrates that our proposed Semantic-OOD and Iter-OOD are effective to make use of the unlabeled seen and unseen class data to improve the OOD detection performances. Secondly, Fig.~\ref{fig4} and Table~\ref{tab9} show that Semantic-OOD outperforms Iter-OOD with a significant margin, especially on CUB and SUN, where the improvements of average TNR are $3.5\%$ and $4.0\%$ respectively. This demonstrates that using semantic information could promote the OOD detection performance. Besides, Table~\ref{tab9} also reports the scale of the simulated unseen-class set in Semantic-OOD and that in Iter-OOD. We find the scale of the simulated unseen-class set in Semantic-OOD is significantly smaller than that of the simulated unseen-class set in Iter-OOD, which indicates that the simulated unseen-class set selected by Semantic-OOD is more effective and representative than that selected by Iter-OOD for the OOD detector training, demonstrating the effectiveness of semantic information for OOD detection.

\begin{table}[t]
	\setlength{\abovecaptionskip}{0cm}	
	\setlength{\belowcaptionskip}{0.1cm}
	\caption{Comparative results of OOD detectors on AWA2, CUB and SUN. aTNR: average TNR, L: the scale of the simulated unseen-class set.}
	\centering
	\begin{tabular}{ccccccc}
		\toprule
		Method& \multicolumn{2}{c}{AWA2}&  \multicolumn{2}{c}{CUB}&   \multicolumn{2}{c}{SUN} \\
		\cmidrule(lr){2-3} \cmidrule(lr){4-5} \cmidrule(lr){6-7}
		& 			       aTNR&	  $L$&		aTNR&	  $L$&			aTNR&	  $L$	  \\
		\cmidrule{1-7}
		MAX-SOFTMAX&  	   0.554&	  -&	      0.234& 	-&	    0.128& 	-\\
		GAN-OD&  	       0.610&	  -&	      0.228& 	-&	    0.144& 	-\\
		Iter-OOD&  	       0.951&	  11000&	  0.506& 	3500&	0.257& 	4000\\
		Semantic-OOD&  	       0.958&	  2444&	  	  0.541& 	508&	0.297& 	270\\
		\bottomrule
	\end{tabular}
	\label{tab9}
\end{table}

\subsection{Discussion}

\subsubsection{Performance on Multi-Model Framework}
Here we evaluate the three 3-model frameworks, i.e. 2A+B, A+2B, A+B+C. Note that the used Model C has slightly worse performance than both Model A and Model B in our current implementation. Specifically, the $ACC$s of Model C are $71.8$, $56.6$, and $59.5$ in AWA2, CUB, and SUN respectively, which are slightly lower than those of Model A and Model B as shown in Table~\ref{tab7}. In each of the three 3-model frameworks, both the cyclic exchanging module (CEM) and the agreement-based exchanging module (AEM) are used respectively. At first, we evaluate the prediction disagreement of the added third base model to the original two base models in the ICoT-ZSL framework by our proposed $APR$ metric. Specifically, we compute the average $APR$ of the added third base model (i.e. Model A, Model B, and Model C in 2A+B, A+2B, A+B+C respectively) to Model A and Model B on AWA2, CUB, and SUN with the PS data split respectively. The results are reported in Table~\ref{tab13}, from which we can see that the A+B+C framework has the largest prediction disagreement among base models. Then, we evaluate the performances of the three 3-model frameworks by conducting conventional ZSL tasks on AWA2, CUB, and SUN respectively. The results are shown in Table~\ref{tab11}. It could be observed from Table~\ref{tab11} and Table~\ref{tab5} that 1) the A+B+C framework, which introduces most complementary information, performs relatively better than the other two 3-model frameworks; and 2) all the three 3-model frameworks achieve close performances to the proposed 2-model ICoT-ZSL (A+B). These observations demonstrate that the number of used base models is not very crucial to the framework's performance, but whether the base models contain complementary information is a relatively more important factor for improving the framework's performance to some extent. This point will be further discussed in the next section. In addition, we note that the two kinds of exchanging modules achieve close performances on the three datasets in most cases. This indicates that both exchanging modules are able to facilitate the exchange of complementary information between the base models.
\begin{table}[t]
	\setlength{\abovecaptionskip}{0cm}	
	\setlength{\belowcaptionskip}{0.1cm}
	\caption{Comparative results ($APR$) of prediction disagreement among base models in the three extended 3-model frameworks on AWA2, CUB, and SUN. Note that the two `A' (or `B') at the two sides of `vs' represent two models with the same architecture but different initialization, and the $APR$ of A vs A,B is computed by averaging the $APR$ of A vs A and that of A vs B, the others are the same.}
	\centering
	\begin{tabular}{cccc}
		\toprule
		Method&          AWA2&  CUB&   SUN \\
		\cmidrule{1-4}
		A vs A,B&  	 12.0&	  21.2&	  17.4\\
		B vs A,B&  	 13.0&	  15.5&	  23.8\\
		C vs A,B&  	 16.2&	  34.0&	  27.1\\
		\bottomrule
	\end{tabular}
	\label{tab13}
\end{table}

\begin{table}[t]
	\setlength{\abovecaptionskip}{0cm}	
	\setlength{\belowcaptionskip}{0.1cm}
	\caption{Results ($ACC$) of the three extended 3-model frameworks on AWA2, CUB, and SUN.}
	\centering
	\begin{tabular}{ccccccc}
		\toprule
		Dataset& \multicolumn{2}{c}{2A+B}& \multicolumn{2}{c}{A+2B}&   \multicolumn{2}{c}{A+B+C} \\
		\cmidrule(lr){2-3} \cmidrule(lr){4-5} \cmidrule(lr){6-7}
			& 	  CEM&   	AEM&		CEM&   	AEM&		CEM&   AEM \\
		\cmidrule{1-7}	
		AWA2&    89.4&	   89.0&       89.7&   89.0&	   90.0&   89.2 \\
		CUB&  	 73.5&     74.4&       75.4&   74.8&       76.5&   76.4 \\
		SUN&	 70.8& 	   70.6&	   70.3&   70.6&	   70.8&   70.4 \\
		\bottomrule
	\end{tabular}
	\label{tab11}
\end{table}

\subsubsection{Effect of Diversity of Model Predictions}
We have pointed out that it is beneficial to the performance of the ICoT-ZSL framework if two base ZSL models have diverse predictions on unseen-class samples. The results in the 3-model cases also to some extent demonstrate the importance of prediction diversity among base models. Here we further empirically demonstrate this claim in the 2-model ICoT-ZSL(A+B) framework by comparing it with the following two 2-model frameworks: 1) A+A using two Model A as base models; 2) B+B using two Model B as base models. Then, we quantitatively measure the prediction disagreement of the two base models used in each framework by computing their $APR$ metrics. Finally, the three frameworks are trained to perform conventional ZSL tasks on AWA2, CUB, and SUN with the PS data split respectively. The results are reported in Table~\ref{tab3}. From Table~\ref{tab3}, we can see that ICoT-ZSL has a larger degree of prediction disagreement than both A+A and B+B on all the three datasets. At the same time, it achieves the best performances on all the three datasets. This demonstrates that the prediction diversity among two base models has a positive effect on the framework' performance. We also note in Table~\ref{tab3} that larger $APR$ does not automatically mean better $ACC$. For instance, B+B has a larger $APR$ than A+A on SUN, however the $ACC$ of B+B is smaller than that of A+A. This is because the performance of the framework is also dependent on the base model's performance, i.e. since Model A has better performance than Model B on SUN, A+A achieves higher accuracy than B+B on SUN, which is in line with our first base-model selection guideline.
\begin{table*}[t]
	\setlength{\abovecaptionskip}{0cm}	
	\setlength{\belowcaptionskip}{0.1cm}
	\caption{Comparative results under different degrees of prediction disagreement among base models in the 2-model frameworks on AWA2, CUB, and SUN.}
	\centering
	\begin{tabular}{ccccccc}
		\toprule
		Method&  \multicolumn{2}{c}{AWA2}& \multicolumn{2}{c}{CUB}&   \multicolumn{2}{c}{SUN} \\
		\cmidrule(lr){2-3} \cmidrule(lr){4-5} \cmidrule(lr){6-7}
		& 	$APR$&   $ACC$&		$APR$&   $ACC$&		$APR$&   $ACC$ \\
		\cmidrule{1-7}
		A+A&   6.8&	 88.0&	10.0&  65.9&	8.4&  69.0\\
		B+B&    9.2&	 87.8&	13.9&  73.1&	21.2&  64.8\\
		ICoT-ZSL(A+B)&   17.3&  89.8&	32.4&  74.6&	26.4&  70.5 \\
		\bottomrule
	\end{tabular}
	\label{tab3}
\end{table*}

\subsubsection{Benefit of the Co-Training Transductive Learning}
Here we investigate the benefit of the iterative co-training transductive learning (CTL) method to ZSL performance. To this end, we perform conventional ZSL tasks on AWA2, CUB and SUN with the PS data split using 1) single Model A without iterative training (denoted as A), 2) two Model A with CTL (denoted as 2A+CTL), 3) single Model B without iterative training (denoted as B), and 4) two Model B with CTL (denoted as 2B+CTL). The results are reported in Table~\ref{tab7}. As seen from Table~\ref{tab7}, 2A+CTL and 2B+CTL significantly outperform A and B on all the three datasets respectively, which demonstrates that iteratively co-training two models with unlabeled unseen-class samples is an effective way to learn more appropriate visual-semantic mappings, hence alleviating the domain shift problem. Besides, we find that the improvements by CTL are relatively small in CUB and SUN. Considering that the performances of the base models on CUB and SUN are relatively low, these indicate that the benefit of CTL depends also on the performance of the base ZSL models. This is also consistent with the our first base-model selection guideline. Intuitively, poor accuracies of base models on the unseen-class samples will result in relatively noisy pseudo labels, thus affecting the iterative co-training of the ZSL models. In addition, we observe that different base ZSL models have different efficiencies with the use of unseen-class samples on specific datasets under CTL. For instance, 2A+CTL improves A by $6.9\%$ while the improvement by CTL on B is only $4.3\%$ when they work on SUN. However, the improvement by CTL on B is larger than that on A when they work on CUB. This observation indicates that co-training two different base ZSL models via an exchanging module could better exploit the potential complementary information from each other.

\begin{figure}
	\centering
	\includegraphics[width=1.0\linewidth]{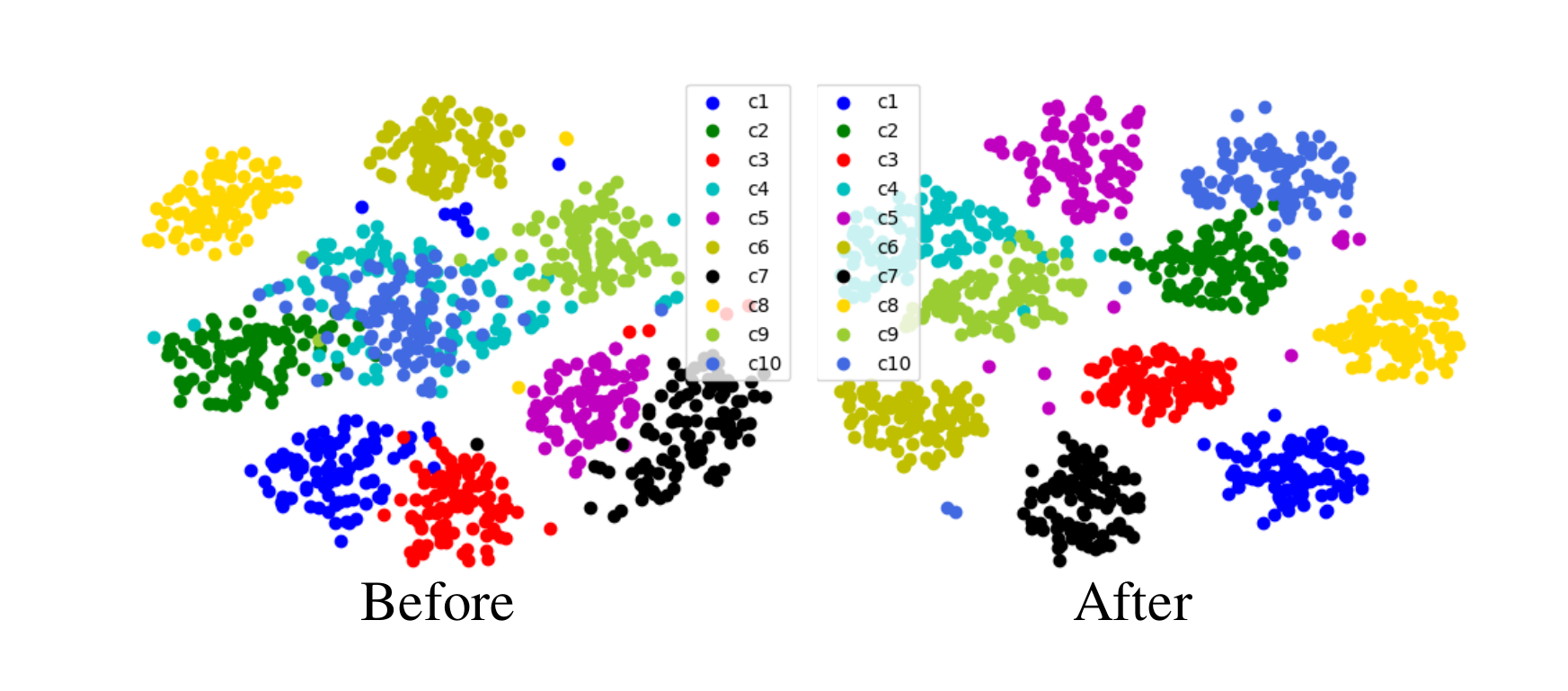}
	\caption{Feature visualization on AWA2. `Before' means visual features before iterative co-training. `After' means visual features after iterative co-training.}
	\label{fig6}
\end{figure}

For a qualitative evaluation, we also visualize the unseen-class features generated by Model B before and after our proposed iterative co-training transductive learning on AWA2, as shown in Fig.~\ref{fig6}. From Fig.~\ref{fig6}, we can see that before the iterative co-training, synthetic features of different unseen classes have a substantial overlap. In contrast, the generated unseen-class features become more distinct after the iterative co-training, which explains to some degree the improved performances by our proposed iterative co-training.
\begin{table}[t]
	\setlength{\abovecaptionskip}{0cm}	
	\setlength{\belowcaptionskip}{0.1cm}
	\caption{Comparative results ($ACC$) of models with/without iterative co-training transductive learning on AWA2, CUB, and SUN.}
	\centering
	\begin{tabular}{cccc}
		\toprule
		Method&          AWA2&  CUB&   SUN \\
		\cmidrule{1-4}
		A&  	 	 72.8&	  57.9&	  62.1\\
		2A+CTL&  	 88.0&	  65.9&	  69.0\\
		B&  	 	 	 72.1&	  57.6&	  60.5\\
		2B+CTL&  	 	 87.8&	  73.1&	  64.8\\
		\bottomrule
	\end{tabular}
	\label{tab7}
\end{table}

\subsubsection{Effect of the Incremental Learning Scheme}
\begin{figure}
	\centering
	\includegraphics[width=0.9\linewidth]{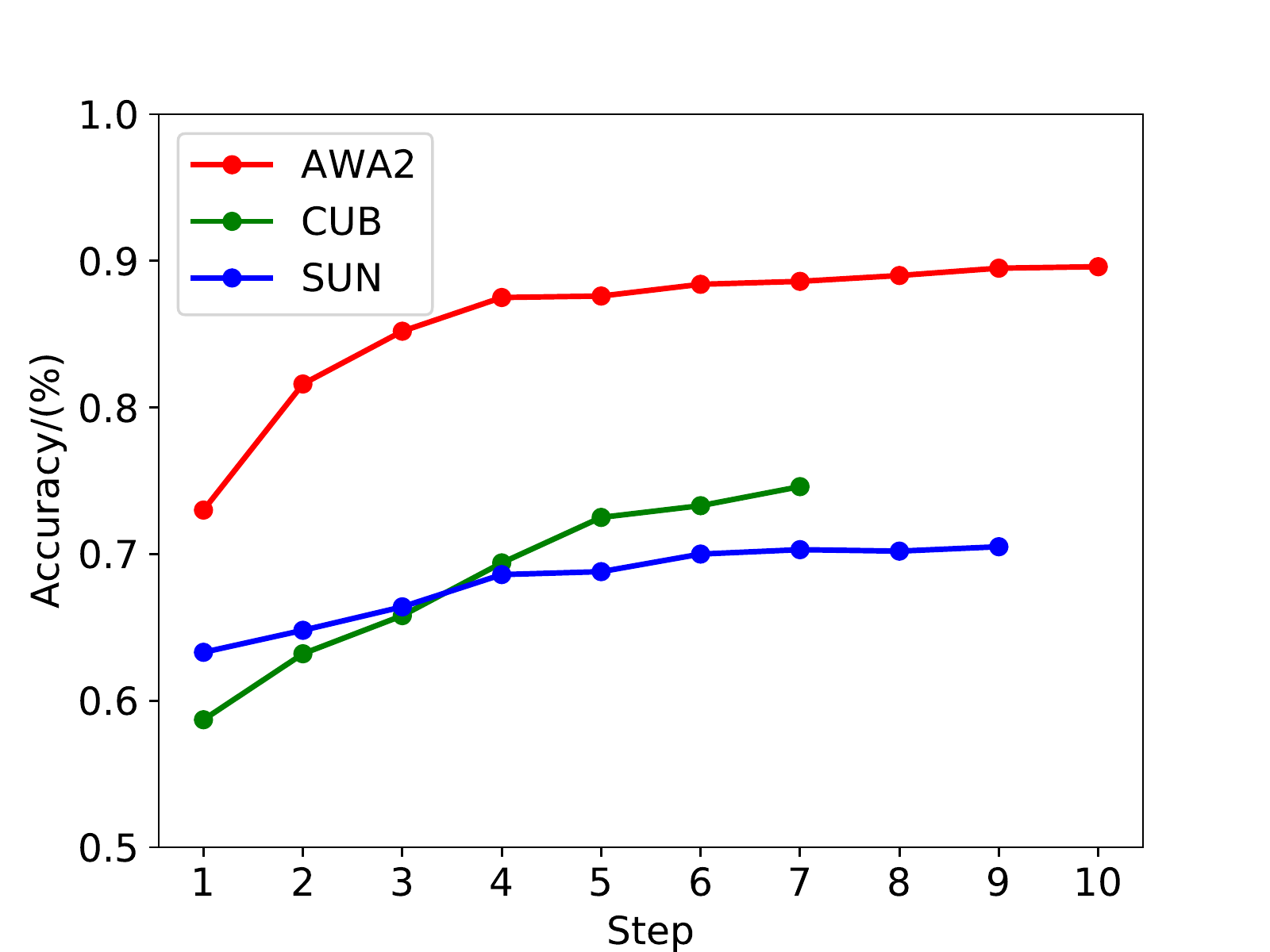}
	\caption{Results ($ACC$) of ICoT-ZSL in the process of incremental learning on AWA2, CUB, and SUN.}
	\label{fig3}
\end{figure}

Here we investigate the effect of the incremental learning scheme on ZSL performance. For comparison, we implement an one-off learning scheme in the proposed ICoT-ZSL framework, where all the unseen-class samples are used for training at every iteration. Note that the difference between the incremental learning scheme and the one-off learning scheme only lies in that the former uses the unseen-class samples in a progressive scheme while the later uses all unseen-class samples at once. The experiments are conducted in the conventional ZSL setting on AWA2, CUB and SUN with the PS data split respectively. The results are reported in Table~\ref{tab8}, which show that the incremental learning scheme is superior over the one-off learning scheme on CUB and SUN while they achieve comparable performances on AWA2. This is because CUB and SUN are two relatively harder benchmarks where the pseudo labels predicted by the base ZSL models at the initial steps are less reliable. If all the pseudo-labeled unseen-class samples are used at once to train the models, the models are more likely to be corrupted by such noisy unseen-class data. While AWA2 is a relatively easier benchmark where the pseudo labels are more accurate, hence the progress to use the unseen-class samples has a relatively small effect on the final ZSL performance. In addition, we also record the results at every iteration in the process of incremental learning, as shown in Fig.~\ref{fig3}. It can be seen from Fig.~\ref{fig3} that the $ACC$ of the proposed ICoT-ZSL method gradually increases as the iterative training goes on and reaches a stable performance finally. This demonstrates that the proposed ICoT-ZSL method is effective to progressively make use of the unlabeled unseen-class samples and promote the ZSL performances incrementally.
\begin{table}[t]
	\setlength{\abovecaptionskip}{0cm}	
	\setlength{\belowcaptionskip}{0.1cm}
	\caption{Comparative results ($ACC$) of incremental learning and one-off learning under the ICoT-ZSL framework on AWA2, CUB, and SUN.}
	\centering
	\begin{tabular}{cccc}
		\toprule
		Method&          AWA2&  CUB&   SUN \\
		\cmidrule{1-4}
		Incremental&  	 89.8&	  74.6&	  70.5\\
		One-off&  	     89.7&	  70.1&	  68.9\\
		\bottomrule
	\end{tabular}
	\label{tab8}
\end{table}

\subsubsection{Effect of Prediction Weight}
\begin{figure}
	\centering
	\includegraphics[width=0.9\linewidth]{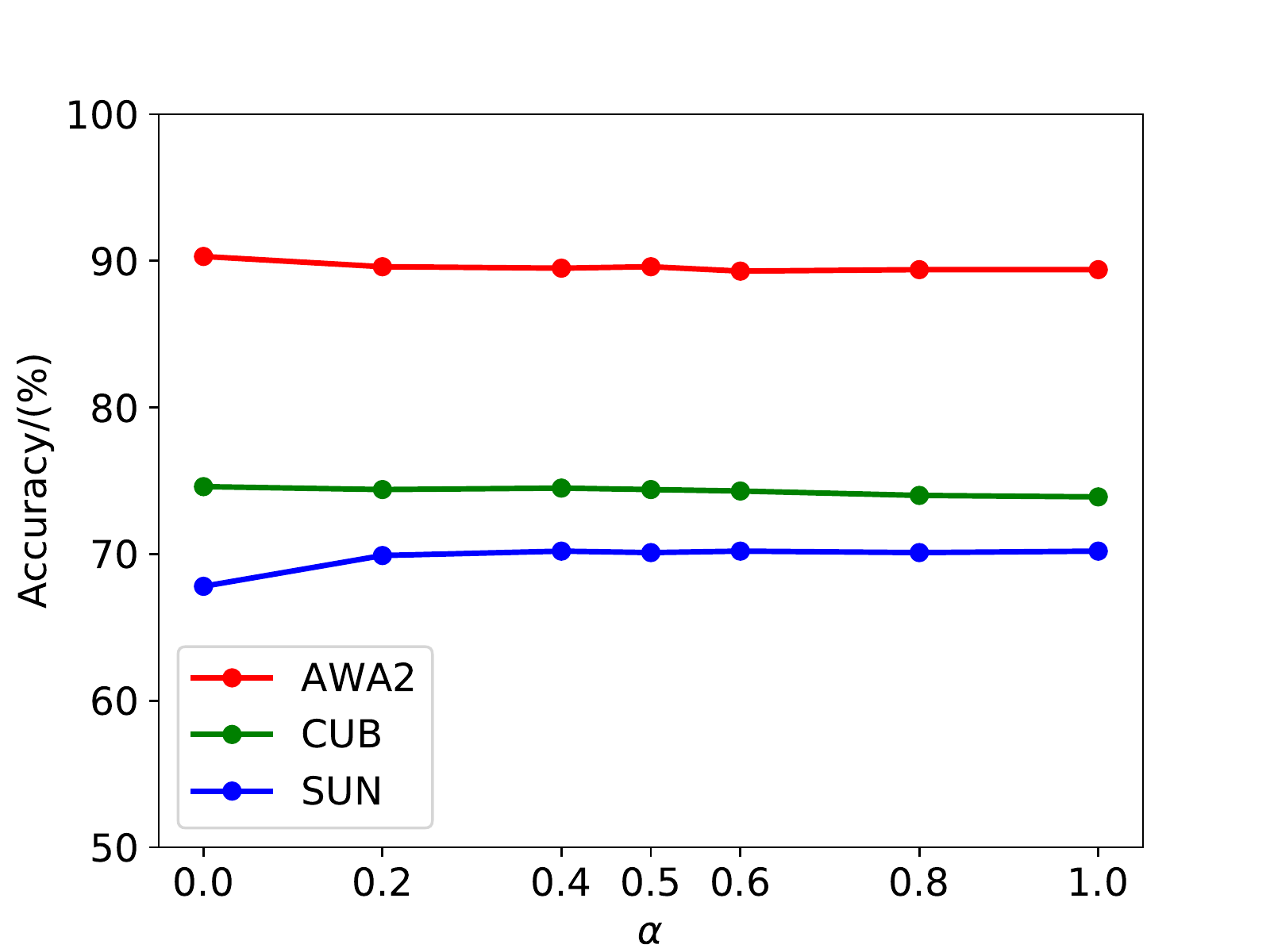}
	\caption{Results ($ACC$) of ICoT-ZSL under different choices of $\alpha$ on AWA2, CUB, and SUN.}
	\label{fig7}
\end{figure}

In the proposed ICoT-ZSL framework, the final prediction for each testing input is made by computing the weighted sum of the probabilistic predictions from Model A and Model B. Here we evaluate the effect of the prediction weight ($\alpha$) on the ICoT-ZSL's performance by conducting ZSL tasks on AWA2, CUB, and SUN with the PS data split respectively, with $\alpha=\{0.0,0.2,0.4,0.5,0.6,0.8,1.0\}$. The results are shown in Fig.~\ref{fig7}. From Fig.~\ref{fig7}, we find that when $\alpha$ is varied from $0.0$ to $1.0$, the $ACC$s of the proposed ICoT-ZSL on all the three datasets change slightly in most cases. This indicates that the performance of the proposed ICoT-ZSL is insensitive to the choice of $\alpha$.

\subsubsection{Training Time Comparison between Co-Training and Self-Training}
Here we compare the training time of the proposed 2-model co-training framework with that of a single-model self-training framework. To this end, we construct three models to conduct ZSL tasks on AWA2, CUB, and SUN with the PS data split respectively, which includes 1) single Model A with iterative self-training (denoted as A+SelfTraining), 2) single Model B with iterative self-training (denoted as B+SelfTraining), 3) a Model A and a Model B with iterative co-training (i.e. our ICoT-ZSL). We compute training time by two time measures. The one is the overall time (OT) which counts the period between the starting step and the best-performance step and the other is the average per-step time (AT) which computes the average per-step time across all steps between the starting step and the best-performance step. Both the 2-model co-training framework and the single-model self-training frameworks are trained on the same machine with the same training parameters. The results are reported in Table~\ref{tab12}. From Table~\ref{tab12}, we observe that the AT and OT of A+SelfTraining are significantly smaller than those of the other two models, mainly because the training time of the base model A is largely smaller than that of the other base model B. In addition, it is noted that the AT of the proposed ICoT-ZSL is larger than that of B+SelfTraining, but its OT is close to or even smaller than that of B+SelfTraining. This is because ICoT-ZSL could reach the best performance with fewer iterative steps by co-training two models.
\begin{table*}[t]
	\setlength{\abovecaptionskip}{0cm}	
	\setlength{\belowcaptionskip}{0.1cm}
	\caption{Training time comparison (Seconds) of ICoT-ZSL and single-model self-training methods on AWA2, CUB, and SUN.}
	\centering
	\begin{tabular}{ccccccc}
		\toprule
		Method&          \multicolumn{2}{c}{AWA2}& \multicolumn{2}{c}{CUB}&   \multicolumn{2}{c}{SUN} \\
		\cmidrule(lr){2-3} \cmidrule(lr){4-5} \cmidrule(lr){6-7}
		& 	AT&   OT&		AT&   OT&		AT&   OT \\
		\cmidrule{1-7}		
		A+SelfTraining&  	 	 	2.7&	13.4&  		19.3&	  135.3&	15.4&	92.2\\
		B+SelfTraining&  	 	 	198.1&	1386.8& 	273.3&	  1913.4&	397.3&  2781.7\\
		ICoT-ZSL&  		216.6&	1299.8&	    286.7&    1719.9&   437.4&	3061.7 \\
		\bottomrule
	\end{tabular}
	\label{tab12}
\end{table*}

\subsubsection{Effect of OOD Detection on GZSL}
In the second T-GZSL data setting, we propose a two-stage method (named as ICoT-ZSL-SOD) to perform GZSL, where we first perform OOD detection with the proposed Semantic-OOD method before class-level classification. Here we analyze the effect of OOD detection on GZSL performance by conducting GZSL tasks on AWA2, CUB, and SUN with the PS data split using 1) the proposed ICoT-ZSL-SOD method and 2) the proposed ICoT-GZSL method (without OOD detection). The results are reported in Table~\ref{tab10}. From Table~\ref{tab10}, we can see that the ICoT-ZSL-SOD method achieves relatively superior performances over ICoT-GZSL, which demonstrates the benefit of the OOD detection for GZSL tasks. The higher performances achieved by the OOD detection based ICoT-ZSL-SOD are mainly due to its better performances on unseen-class samples since a part of unseen-class samples are picked out by the OOD detector and then classified by a specialized ZSL model. We also note that ICoT-ZSL-SOD is slightly worse than ICoT-GZSL on seen classes on AWA2 and SUN. This is caused by the fact that in order to improve the harmonic mean accuracy, ICoT-ZSL-SOD will increase the unseen-class accuracy by classifying more samples (including real unseen-class samples and a few seen-class samples) as unseen-class ones at the OOD detection stage. As a result, the seen-class accuracy is slightly reduced since a few real seen-class samples are wrongly classified as unseen-class ones. Besides, note that the proposed ICoT-GZSL also achieves considerably higher performances even compared with the recent state-of-the-art methods shown in Table\ref{tab6}. This indicates that our proposed ICoT-ZSL framework is also fit to the GZSL setting.
\begin{table*}[t]
	\setlength{\abovecaptionskip}{0cm}	
	\setlength{\belowcaptionskip}{0.1cm}
	\caption{Comparative results of ICoT-ZSL-SOD (w OOD) and ICoT-GZSL (w/o OOD) on AWA2, CUB, and SUN. U: $ACC$ on unseen classes, S: $ACC$ on seen classes, and $H$: the harmonic mean of U and S.}
	\centering
	\begin{tabular}{cccccccccc}
		\toprule
		Method& \multicolumn{3}{c}{AWA2}&  \multicolumn{3}{c}{CUB}&   \multicolumn{3}{c}{SUN} \\
		\cmidrule(lr){2-4} \cmidrule(lr){5-7} \cmidrule(lr){8-10}
		& 			       U&	  S&	 H&		  U&	  S&	 H&		U&	  S&	 H	  \\
		\cmidrule{1-10}
		ICoT-GZSL (w/o OOD)&  	       81.5&  82.6&  82.0&	  67.5&  62.0&	64.6& 	47.2&  36.8&	41.3\\
		ICoT-ZSL-SOD (w OOD)&  	   84.8&  81.8&	 83.3& 	  66.6&	 64.1& 	65.4& 	50.4&  36.6&	42.4\\
		\bottomrule
	\end{tabular}
	\label{tab10}
\end{table*}

\subsubsection{Sensitivity of OOD Detector Training to the Scale of Simulated Unseen-Class Set}
\begin{figure}
	\centering
	\includegraphics[width=0.9\linewidth]{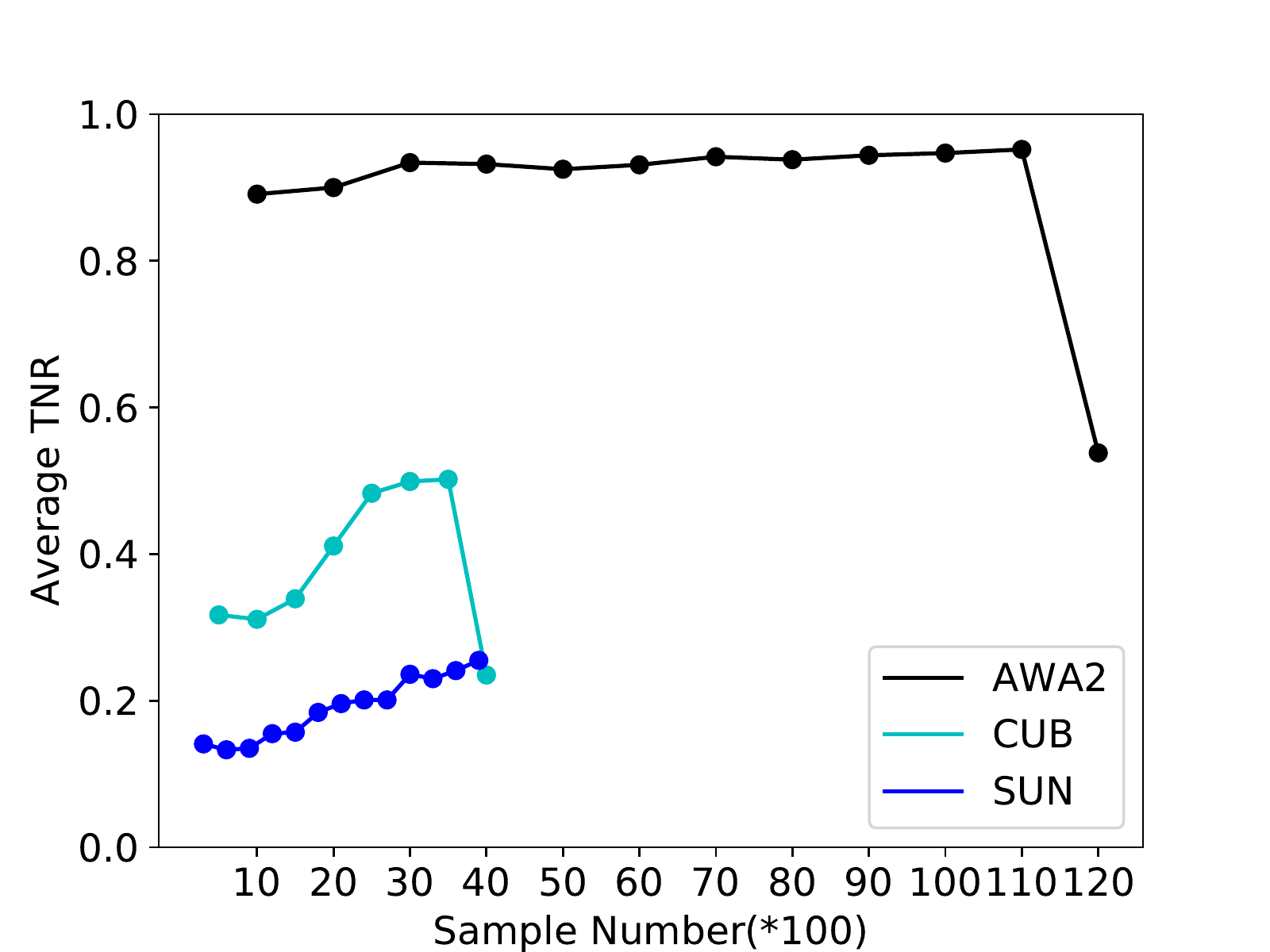}
	\caption{Performances of OOD detectors trained with simulated unseen-class sets with different scales on AWA2, CUB, SUN.}
	\label{fig5}
\end{figure}

Here we analyze the sensitivity of OOD detector training to the scale of the simulated unseen-class set. We conduct this investigation using the Iter-OOD method. In the Iter-OOD method, a number ($L$) of samples with the lowest confidences are firstly selected by a base OOD detector, then these samples are used as the simulated unseen-class set to re-train a new OOD detector. Limited by the accuracy of the base OOD detector, the simulated unseen-class set includes not only unseen-class sampels, but also several seen-class samples which are harmful to the training of the new OOD detector. Hence, the scale of the simulated unseen-class set is a crucial hyper-parameter. To assess the sensitivity of OOD detector training to this hyper-parameter, we vary the scale (i.e. $L$) of the simulated unseen-class set and train several OOD detectors with these simulated unseen-class sets, and evaluate their performances respectively. The experiments are conducted on AWA2, CUB, and SUN with the PS data split respectively. The results are reported in Fig.~\ref{fig5}. As seen from the curves in Fig.~\ref{fig5}, at the initial stages, the performance increases as $L$ becomes larger on all the three datasets. This is because more real unseen-class samples are selected to train the OOD detector as $L$ increases. When $L$ reaches a certain scale, the OOD detection performance stabilizes. Fig.~\ref{fig5} shows that the stable stage has a relatively wide range, which demonstrates that the OOD detection performances are not too sensitive to the scale of the simulated unseen-class set within a relatively large extent. However, if we select a very large simulated unseen-class set, we have the risk of learning a poor OOD detector since too many seen-class samples are contained in the simulated unseen-class set.

\section{Conclusion and Future Work}
\label{conclusion}
In this paper, we propose a general iterative co-training transductive framework for ZSL (called ICoT-ZSL) to effectively make use of the unlabeled unseen-class data to alleviate the domain shift problem. In the ICoT-ZSL framework, two base ZSL models are iteratively co-trained to learn their visual-semantic mappings simultaneously by fully exploiting the complementarity of their classification capabilities. In addition, we also adapt the proposed ICoT-ZSL framework to GZSL. To alleviate the bias problem in GZSL, we propose a semantic-guided OOD detection method to pick out unseen-class samples from the compound unseen-class and seen-class samples before class-level classification. Combined with semantic-guided OOD detector, the adapted ICoT-ZSL framework is effective for GZSL. We also demonstrate that our proposed framework could be easily extended to the multi-model cases by constructing three 3-model frameworks. Extensive experimental results on three benchmarks with two data splits demonstrate that the proposed methods could significantly outperform many state-of-the-art methods with large margins. It is noted that the proposed ICoT-ZSL framework could freely accommodate any existing inductive ZSL models to improve their performances. Furthermore, our proposed ICoT-ZSL framework is a general framework which could be easily adapted to other zero-shot tasks such as multi-label zero-shot learning (ML-ZSL) and zero-shot object detection (ZSD) simply by 1) employing two ML-ZSL (or ZSD) base models to predict pseudo labels for unlabeled unseen-class samples and 2) adding an exchanging module to exchange pseudo labels between two base models for model re-training. As a future work, we will apply the proposed ICoT-ZSL framework to the ML-ZSL and ZSD settings. Besides, in our current implementation, the selection of unseen-class subsets is accomplished by a random sampling method. Considering that different unseen-class samples probably play different roles for the training of ZSL models, in the future, we could exploit other more suitable ways to select the unseen-class subsets to further boost the ZSL performance.


\section*{Acknowledgment}
This work was supported by the National Natural Science Foundation of China (NSFC) under Grants (61991423, U1805264) and the Strategic Priority Research Program of the Chinese Academy of Sciences (XDB32050100).

\ifCLASSOPTIONcaptionsoff
  \newpage
\fi

\bibliographystyle{IEEEtran}
\bibliography{ref}

\end{document}